\documentclass[preprint,10pt]{elsarticle}
\usepackage[letterpaper,top=2.75cm,bottom=2.75cm,left=2.75cm,right=2.75cm,marginparwidth=2.75cm]{geometry}

\usepackage[T1]{fontenc}
\usepackage[utf8]{inputenc} 
\usepackage{microtype}

\usepackage[letterpaper,top=2.75cm,bottom=2.75cm,left=2.75cm,right=2.75cm,marginparwidth=2.75cm]{geometry}

\usepackage{mathtools}  
\usepackage{amsthm,amsfonts,bm}

\usepackage{graphicx,booktabs,multirow,dcolumn,subcaption}
\usepackage{algorithm,algpseudocode}
\usepackage{listings}

\biboptions{numbers,sort&compress}

\usepackage[backref=page]{hyperref}
\usepackage[nameinlink,capitalize]{cleveref}
\Crefname{equation}{Eq.}{Eqs.}
\crefname{equation}{Eq.}{Eqs.}

\usepackage{xcolor}

\Crefname{equation}{Eq.}{Eqs.}

\journal{Elsevier}

\begin{document}

\begin{frontmatter}

\title{Physically Interpretable Representation Learning with Gaussian Mixture Variational AutoEncoder (GM-VAE)}


\author[1]{Tiffany Fan}
\ead{tiffan@stanford.edu}
\author[2]{Murray Cutforth}
\ead{mcc4@stanford.edu}
\author[1,3]{Marta D'Elia\corref{mycorrespondingauthor}}
\cortext[mycorrespondingauthor]{Corresponding author}
\ead{marta@atomicmachines.com}
\author[4]{Alexandre Cortiella}
\ead{alexandre.cortiella@gmail.com}
\author[5]{Alireza Doostan}
\ead{doostan@colorado.edu}
\author[1,2]{Eric Darve}
\ead{darve@stanford.edu}
\affiliation[1]{address={Institute for Computational and Mathematical Engineering, Stanford University}}
\affiliation[2]{address={Department of Mechanical Engineering, Stanford University}}

\affiliation[3]{address={Atomic Machines}}
\affiliation[4]{address={National Renewable Energy Laboratory}}
\affiliation[5]{address={Ann and H.J. Smead Department of Aerospace Engineering Sciences, University of Colorado, Boulder}}

\begin{abstract}
Extracting compact, physically interpretable representations from high-dimensional scientific data is a persistent challenge due to the complex, nonlinear structures inherent in physical systems. We propose a Gaussian Mixture Variational Autoencoder (GM-VAE) framework designed to address this by integrating an Expectation-Maximization (EM)-inspired training scheme with a novel spectral interpretability metric. Unlike conventional VAEs that jointly optimize reconstruction and clustering (often leading to training instability), our method utilizes a block-coordinate descent strategy, alternating between expectation and maximization steps. This approach stabilizes training and naturally aligns latent clusters with distinct physical regimes. To objectively evaluate the learned representations, we introduce a quantitative metric based on graph-Laplacian smoothness, which measures the coherence of physical quantities across the latent manifold. We demonstrate the efficacy of this framework on datasets of increasing complexity: surface reaction ODEs, Navier-Stokes wake flows, and experimental laser-induced combustion Schlieren images. The results show that our GM-VAE yields smooth, physically consistent manifolds and accurate regime clustering, offering a robust data-driven tool for interpreting turbulent and reactive flow systems.
\end{abstract}
\end{frontmatter}

\section{Introduction}
\label{sec:introduction}

High-fidelity analysis of complex, transient, or multiscale flows generates vast quantities of high-dimensional data that are increasingly difficult to interpret. This challenge is particularly acute in experimental diagnostics such as Schlieren imaging, which visualizes refractive index gradients to capture turbulence, shock waves, and combustion dynamics~\cite{settles2001schlieren}. While invaluable for model validation, large-volume Schlieren data present unique analytical hurdles: the images are high-dimensional, sensitive to experimental noise, and exhibit a complex, nonlinear coupling between optical contrast and the underlying thermofluid physics~\cite{echekki2010turbulent,giusti2019turbulent}. Consequently, extracting compact, physically interpretable representations from such datasets remains a critical open problem.

These challenges highlight the need for data-driven frameworks that can extract physically meaningful low-dimensional patterns from high-dimensional flow fields. In this work, we focus on analyzing and interpreting high-dimensional data generated by CFD simulations and Schlieren experiments, seeking representations that reveal the underlying structure and dynamics of the flow in a compact, interpretable form.

Dimensionality reduction approaches are critical for understanding high-dimensional datasets such as Schlieren images. State-of-the-art methods, such as t-SNE~\cite{van2008visualizing}, UMAP~\cite{mcinnes2018umap}, LLE~\cite{roweis2000nonlinear}, and Isomap~\cite{tenenbaum2000global}, focus on preserving the local topology of the original dataset but often distort global structures. Moreover, these methods typically rely on Euclidean ($L_2$) distances to measure neighborhood relationships, which become less informative in high-dimensional spaces where pairwise distances tend to concentrate and lose contrast~\cite{aggarwal2001surprising}. To address these challenges in interpretable dimension reduction of scientific data, this work aims to encode high-dimensional data into physically meaningful, low-dimensional representations that capture key features, preserve both global and local structures, and facilitate interpretation and analysis within the learned latent manifold.

While manifold learning techniques (t-SNE, UMAP) offer visualization, they lack a generative mechanism to map back to the physical space. Conversely, standard Variational Autoencoders (VAEs) provide generative capabilities but typically enforce a single Gaussian prior, which over-smooths the latent space and fails to separate distinct physical regimes (e.g., laminar vs. turbulent). To address this, we employ the Gaussian Mixture Variational Autoencoder (GM-VAE)~\cite{jiang2016variational,fan2025physically} and propose an EM-based training strategy to analyze a novel and complex dataset of Schlieren images generated from real laser-induced combustion systems. Traditional nonlinear dimensionality reduction methods have proven inadequate in capturing the intricate dynamics of this dataset, as demonstrated in this work. The GM-VAE provides a unified probabilistic framework for representation learning, enabling the extraction of low-dimensional latent coordinates that organize the data according to physically meaningful states. These latent variables naturally cluster images corresponding to distinct combustion regimes and allow interpolation between them, revealing smooth transitions in flow evolution.

To further evaluate and interpret the learned latent representations, we introduce a quantitative metric for physical interpretability based on the smoothness of physical quantities across the latent manifold. Because the underlying physical system evolves through smooth transitions in quantities such as pressure and temperature, a physically consistent latent representation should exhibit similar smooth variations of these quantities across the latent space. We formalize this notion using spectral graph theory. A graph is constructed from latent point clouds, where edges are defined between nearest neighbors. The eigenvectors of the graph Laplacian define frequency components on the manifold, allowing any physical variable to be viewed as a signal on this graph. The smoothness of a signal, quantified by the concentration of its energy in low-frequency modes, thus serves as a proxy for interpretability.

To summarize, our main contributions are threefold:
\begin{itemize} 
    \item Robust Training Framework: We propose an EM-inspired block-coordinate descent algorithm for training GM-VAEs. By alternating between latent cluster assignment and parameter updates, this approach stabilizes optimization and ensures that latent clusters align with distinct physical regimes. 
    \item Spectral Interpretability Metric: We introduce a quantitative metric based on graph-Laplacian smoothness to evaluate the physical consistency of learned representations. This metric objectively measures how coherent physical state variables are across the latent manifold. 
    \item Validation: We demonstrate the efficacy of our framework on datasets of increasing complexity: low-dimensional surface reaction ODEs, chaotic Navier-Stokes wake flows, and high-dimensional Schlieren images from laser-induced combustion experiments. 
\end{itemize}

\section{Related Work}
\label{sec:related-work}

\paragraph{Data-Driven Methods for Turbulence and Combustion}
Machine learning has increasingly been adopted to accelerate and enrich turbulent combustion modeling. Ihme et al.~\cite{ihme2022combustion} review data-centric strategies for constructing reduced kinetic mechanisms and low-dimensional manifolds, highlighting the role of data-driven surrogates in complex reacting flows. Complementing this perspective, Duraisamy et al.~\cite{duraisamy2019turbulence} survey physics-informed ML approaches for turbulence modeling, emphasizing closure learning, model-form uncertainty quantification, and the integration of data with governing physical constraints. 
Despite these advances, ensuring that latent representations are robustly correlated with known physical quantities, such as temperature, pressure, species concentration, and reaction rates, remains a challenging task.

\paragraph{Interpretable Representation Learning for Scientific Data}
Interpretable representation learning is essential in scientific domains, where low-dimensional embeddings must be physically meaningful. Variational Autoencoders (VAEs)~\cite{kingma2013auto} compress data into a latent space, but standard VAEs often produce entangled factors. The $\beta$-VAE~\cite{higgins2017beta} improves upon this by promoting disentanglement, while Gaussian Mixture  Variational Autoencoders (GM-VAEs)~\cite{dilokthanakul2016deep} extend the VAE framework by combining clustering with representation learning. GM-VAEs have found applications in various fields. For example, Brunton and Kutz~\cite{brunton2022data} demonstrate how they can reveal bifurcation structures, and Wang et al.~\cite{wang2022deep} utilize them for enhanced reconstruction in Schlieren diagnostics. Our work extends the Variational Deep Embedding framework of Jiang et al.~\cite{jiang2016variational}, which unified deep generative modeling and clustering through a Gaussian-mixture prior in the latent space. Unlike their joint optimization of reconstruction and cluster assignment within a single variational objective, our approach introduces an EM-inspired alternating training strategy that explicitly separates the expectation (cluster-responsibility estimation) and maximization (decoder and encoder update) steps. This block-descent procedure enhances convergence stability and ensures that latent clusters correspond to physically distinct operating regimes, thereby combining the interpretability of physics-informed representations with the flexibility of mixture-based generative models.

Recent conceptual analyses of interpretability in scientific machine learning, such as~\citet{rowan2025interpretability}, have argued that much of the existing SciML literature—particularly in equation discovery and symbolic regression—tends to equate interpretability with mathematical sparsity (see, e.g.,~\citealp{brunton2016sindy, champion2019data, udrescu2020aiFeynman}). However, as~\citet{rowan2025interpretability} emphasize, sparsity alone does not ensure scientific understanding: genuine interpretability requires a mechanistic connection between a model’s learned relationships and the underlying physical principles that govern a system. This position aligns with broader discussions on interpretable machine learning that emphasize understanding mechanisms rather than merely achieving transparency or post-hoc explainability~\citep{rudin2021interpretable, murdoch2019definitions}. Within this perspective, interpretability in SciML is achieved when learned structures can be derived from, or consistently related to, fundamental conservation laws or constitutive mechanisms, rather than simply taking compact analytic forms.

Our GM-VAE framework adopts this mechanism-based definition: instead of enforcing sparsity in network parameters, we evaluate interpretability through spectral-graph metrics that quantify the smoothness of latent manifolds with respect to physically meaningful quantities, such as \(g^*\), Reynolds number, and ignition regime. This smoothness-based criterion formalizes the physical expectation that continuous variables—such as pressure, temperature, and species concentration—should vary continuously across a well-structured latent space. If two states are physically similar, their latent representations should be close, and the associated physical quantities should not exhibit high-frequency oscillations on the latent manifold. Conversely, if the latent coordinates separate physically similar states, the induced physical signals become non-smooth, indicating that the representation does not preserve physical locality. In this sense, smoothness provides an operational proxy for mechanistic interpretability, following the perspective of~\citet{rowan2025interpretability}, because it measures whether the learned coordinates respect the continuity inherent in the governing physical processes. A conceptually related perspective appears in the materials science literature. For example,~\citet{desai2024trade} evaluates the physical plausibility of latent trajectories based on their temporal smoothness, arguing that smoothly varying latent coordinates more faithfully capture microstructural evolution. Although their notion of smoothness is defined along time rather than across a data-driven manifold, the underlying principle is analogous: smooth latent variation acts as a proxy for physically consistent embedding.

\paragraph{Graph-Based Smoothness Metrics and Spectral Methods}
Graph-based techniques provide a natural framework for quantifying the smoothness of functions defined on irregular point clouds. Belkin and Niyogi~\cite{belkin2003laplacian} introduce Laplacian Eigenmaps for manifold learning, where the eigenfunctions of the graph Laplacian serve as smooth basis functions. Von Luxburg~\cite{von2007tutorial} reviews how spectral graph theory underpins techniques such as diffusion maps and spectral clustering. In thermal field reconstruction from sparse sensors, Ross et al.~\cite{ross2022projecting} show that graph-Laplacian-based propagation effectively captures coherent, large-scale thermal structures associated with low-frequency graph modes. This observation is consistent with graph signal processing theory, which establishes that low graph frequencies correspond to smooth, global variations over the underlying manifold~\cite{shuman2013emerging}. More recently, Bronstein et al.~\cite{bronstein2017geometric} survey geometric deep learning approaches, including graph convolutional networks, that inherently exploit graph-Laplacian smoothness priors. Our proposed interpretability metric builds on this foundation: we treat Schlieren-derived latent embeddings as nodes of a graph, compute Laplacian eigenvectors, and measure how physical quantities (e.g., pressure, temperature) project onto the resulting spectral basis. A quantity that concentrates most of its spectral energy in low-frequency graph eigenmodes varies smoothly across the latent manifold, indicating that the latent geometry preserves underlying physical continuity. If the spectral energy instead spreads across high-frequency modes, the corresponding physical quantity varies rapidly and inconsistently across the manifold, signaling that physically similar states are placed far apart in latent space. Thus, smoothness in the graph-spectral sense provides a principled, quantitative measure of physical interpretability: it tests whether the learned embedding organizes flow fields in a manner consistent with their mechanistic structure.

\section{Methodology}
\label{sec:method}

This section details the methodology employed in our study, encompassing the architecture and training process of the GM-VAE.

\subsection{Model Assumptions}
\label{subsec:model_assumptions}

We employ a Gaussian Mixture Variational Autoencoder (GM-VAE) as our primary model for dimensionality reduction, clustering, and generative modeling. The core assumption is that the high-dimensional scientific data considered in this work---including reaction trajectories, wake-flow snapshots, and Schlieren images---lie on a low-dimensional manifold whose structure reflects distinct physical regimes. These regimes may correspond, for example, to different ignition outcomes, pressure levels, laser locations, or flow topologies. The latent clusters in the GM-VAE are therefore intended to group data according to such underlying physical states~\cite{fan2025physically}.

\begin{figure}[htbp]
    \centering
    \includegraphics[width=0.95\linewidth]{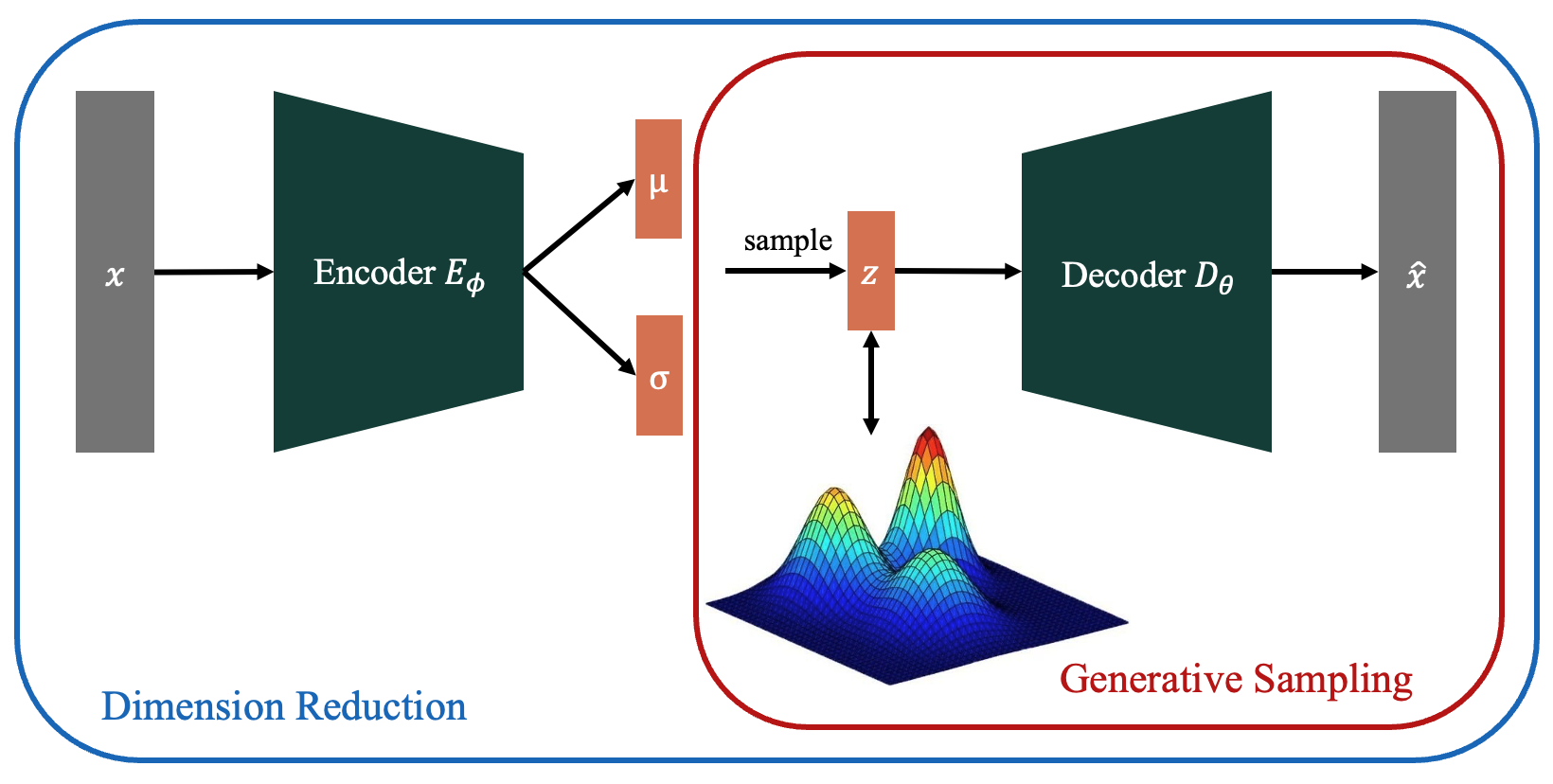}
    \caption{The GM-VAE framework enables both dimension reduction and generative sampling. In addition to the standard VAE pipeline, this framework introduces a Gaussian Mixture prior for the latent variables to impose a clustering structure in the latent space, which reflects distinct physical regimes.
Conceptually, the GM-VAE assumes that our data naturally form clusters corresponding to distinct physical regimes or states. During training, the model jointly minimizes reconstruction error and encourages the latent representations to align with a Gaussian-mixture distribution. The model is thus encouraged to learn a latent manifold where nearby points share similar physical characteristics, and transitions between clusters reflect real physical transitions.
}
    \label{fig:model}
\end{figure}

The GM-VAE pipeline, illustrated in \Cref{fig:model}, integrates a standard Variational Autoencoder (VAE) with a Gaussian Mixture Model (GMM) imposed on the latent space. In the original GM-VAE formulation~\cite{jiang2016variational}, the encoder, decoder, and mixture components are trained jointly by maximizing a single variational objective. In contrast, the present work adopts an expectation--maximization (EM)-inspired training procedure: the encoder and decoder parameters are updated in the maximization step, while cluster responsibilities and mixture statistics are updated in the expectation step. This alternating scheme reduces competition between reconstruction and clustering objectives, stabilizes training, and encourages the latent variables to align more clearly with meaningful physical regimes. A key advantage of the GM-VAE over a standard VAE is its ability to impose explicit cluster structure in the latent space. Whereas the VAE typically produces an entangled latent representation that mixes data from different physical regimes, the GM-VAE assigns each latent point a probabilistic cluster label, encourages separation of regimes, and enables sampling or interpolation within a specific physical state. This structured latent space provides a natural foundation for the graph-spectral interpretability metric introduced in \Cref{sec:evaluation_metric}.

Mathematically, we assume that the \( N \) training examples are independently and identically distributed (i.i.d.) and follow a Gaussian Mixture Model in the latent space:
\begin{align*}
    c & \sim \text{Cat}(K, \pi), \\
    z \mid c & \sim \mathcal{N}(\mu_c, \sigma_c^2 I), \\
    x \mid z & \sim \mathcal{N}(\tilde{\mu}, \tilde{\sigma}^2 I),
\end{align*}
where:
\begin{itemize}
    \item \( c \) is a categorically distributed random variable denoting the cluster label.
    \item \( K \) is the number of clusters (a chosen hyperparameter).
    \item \( \pi \in \mathbb{R}_+^K \) is the prior probability vector for the categorical distribution of class label \( c \), satisfying \( \sum_{k=1}^K \pi_k = 1 \).
    \item \( \mu_c \) and \( \sigma_c^2 \) denote the mean and variance of cluster \( c \), respectively.
    \item \( \tilde{\mu} \) and \( \tilde{\sigma}^2 \) represent the conditional mean and variance of the data.
    \item \( I \) is the identity matrix.
\end{itemize}

In all examples, the number of latent clusters \(K\) is selected to match the expected number of distinct physical regimes present in the data. For the reaction-kinetics and wake-flow cases, these regimes correspond to qualitatively different dynamical behaviors (e.g., distinct vortex-shedding patterns). For the Schlieren datasets, \(K\) is chosen empirically based on known experimental conditions such as ignition outcome, pressure level, laser-spark location, as well as the resulting latent embedding distribution. Because the GM-VAE assigns probabilistic cluster memberships, the method remains robust even if regimes are not perfectly separated, but choosing \(K\) consistent with known physics improves stability of the EM updates and interpretability of the resulting latent manifold.

\subsection{Loss Function}
\label{subsec:loss_function}

The training objective of the GM-VAE is to maximize the Evidence Lower Bound (ELBO) of the log-likelihood of the data~\cite{kingma2013auto}. The ELBO serves as a surrogate for the intractable log-likelihood, enabling efficient optimization through variational inference.

\paragraph{Joint Probability and ELBO}

For a single training example \( x \), the joint probability with latent variables \( z \) and cluster label \( c \) is given by:
\[
p(x, z, c) = p(c) \, p(z \mid c) \, p(x \mid z).
\]
Applying Jensen's inequality, we derive the ELBO as follows:
\begin{align*}
    \log p(x) &= \log \int_z \sum_c p(x, z, c) \, dz \\
    &\geq \mathbb{E}_{q(z, c \mid x)} \left[ \log \frac{p(x, z, c)}{q(z, c \mid x)} \right] = \mathcal{L}_{\text{ELBO}}(x),
\end{align*}
where \( q(z, c \mid x) \) is the variational posterior distribution.

\paragraph{Mean-Field Approximation}

To simplify the computation, we adopt the mean-field approximation, assuming that the variational posterior factorizes as
\[
q(z, c \mid x) = q^{(z)}(z \mid x) \, q^{(c)}(c \mid x).
\]
Following~\cite{trask2022unsupervised, jiang2016variational}, we approximate the posterior probability as
\[
q^{(c)}(c \mid x) \approx p(c \mid z) = \frac{p(c) \, p(z \mid c)}{\sum_{c'} p(c') \, p(z \mid c')} := \gamma_c.
\]
Here, \( q^{(z)}(z \mid x) \) is modeled as a Gaussian distribution with mean \(\mu \) and variance \( \sigma^2 \), parameterized by the encoder neural network.

\paragraph{ELBO Expression}

Under these assumptions, the ELBO for a single example \( x \) can be expressed as
\begin{align*}
    \mathcal{L}_{\text{ELBO}}(x) &= \mathbb{E}_{q(z \mid x)} \left[ \log p(x \mid z) \right] 
    + \mathbb{E}_{q(z, c \mid x)} \left[ \log p(z \mid c) \right] 
    + \mathbb{E}_{q(c \mid x)} \left[ \log p(c) \right] \\
    &\quad - \mathbb{E}_{q(z \mid x)} \left[ \log q^{(z)}(z \mid x) \right] 
    - \mathbb{E}_{q(c \mid x)} \left[ \log q^{(c)}(c \mid x) \right].
\end{align*}
This formulation allows us to decompose the ELBO into reconstruction and regularization terms, facilitating efficient optimization using gradient-based methods.

\paragraph{Reparameterization Trick}

To enable gradient back-propagation through the sampling process, we employ the reparameterization trick,
\[
\epsilon^{(n)} \sim \mathcal{N}(0, I), 
\quad 
z^{(n)} = \mu + \sigma \circ \epsilon^{(n)}
\]
where \( \circ \) denotes element-wise multiplication, and \( n \) indexes the training set.

\subsection{Training Strategy}
\label{subsec:training_strategies}

In the GM-VAE, the latent variables are assumed to arise from a mixture of $K$ Gaussian components, each representing a distinct physical regime. During training, the mixture parameters, i.e., cluster weights, means, and variances, must be updated to reflect the evolving structure of the latent representations produced by the encoder.  

The expectation--maximization (EM) viewpoint provides a natural way to interpret this update. In the E-step, the encoder produces a latent point $z^{(n)}$ for each data sample. Each latent point is then assigned a soft responsibility for belonging to each cluster. These responsibilities, denoted $\gamma_c^{(n)}$, quantify the probability that $z^{(n)}$ originated from cluster $c$.  

In the M-step, the mixture parameters are updated by treating these responsibilities as fractional cluster memberships. The cluster weights become the average responsibilities, the cluster means become responsibility-weighted averages of latent points, and the variances reflect the spread of latent points around each mean.  

Interleaving these EM-style updates with gradient-based optimization of the encoder and decoder helps stabilize training: the responsibilities adapt to the evolving latent space, while the latent space is simultaneously encouraged to form well-separated, coherent regions corresponding to physically meaningful regimes. In addition, we introduce a $\beta$-weighted KL regularization term that softly pulls the latent representations toward a neighborhood of the standard normal prior, preventing pathological cluster collapse while maintaining a smooth, well-conditioned manifold.

We summarize the complete training procedure in Algorithm~\ref{alg:basic}.

\begin{algorithm}[htbp]
\caption{GM-VAE Training with EM Algorithm}\label{alg:basic}
\begin{algorithmic} 
\State \textbf{Input}: Data \( \{ x^{(n)} \} \), \( n=1, \ldots, N \) with mini-batch index sets $B_b$ , \( b=1, \ldots, N_B \) 
\While{not converged}
    \For{each training mini-batch \( \{ x^{(n)} \}_{n\in B_b} \)} \Comment{Update VAE via Gradient Descent}
        \State \textbf{Encode}: \( \mu^{(n)}, \sigma^{2(n)} \gets E_{\phi}(x^{(n)}), \, \forall n\in B_b \), where $E_{\phi}$ is the encoder network parameterized by $\phi$ and $\mu^{(n)}, \sigma^{2(n)}$ specify a Gaussian distribution for the latent embedding for the $n$-th data point
        \State \textbf{Sample}: \( z^{(n)} \sim \mathcal{N}(\mu^{(n)}, \sigma^{2(n)} I),\, \forall n\in B_b \)
        
        \State \textbf{Decode}: \( \tilde{\mu}^{(n)}, \tilde{\sigma}^{2(n)} \gets D_{\theta}(z^{(n)}), \, \forall n\in B_b \), where $D_{\theta}$ is the decoder network with parameters $\theta$ and $\tilde{\mu}^{(n)}, \tilde{\sigma}^{2(n)} $ specify a Gaussian distribution for reconstructing the $n$-th data point
        \State \textbf{Compute ELBO}: 
        \begin{align*}
            \mathcal{L}^{(n)}_{\text{ELBO}} 
                &= \log \mathcal{N}\!\bigl(x^{(n)} \mid \tilde{\mu}^{(n)}, \tilde{\sigma}^{2(n)}\bigr) \\
                &\quad - \frac{1}{2} \sum_c \gamma_c^{(n)} \sum_{j} \left( 
                    \log \bigl(2\pi \sigma_{c,j}^{2}\bigr) 
                    + \frac{\sigma_j^{2(n)}}{\sigma_{c,j}^{2}} 
                    + \frac{\bigl(\mu_j^{(n)} - \mu_{c,j}\bigr)^2}{\sigma_{c,j}^{2}} \right) \\
                &\quad + \frac{1}{2} \sum_{j} \left( \log \bigl(2\pi \sigma_j^{2(n)}\bigr) + 1 \right) 
                    + \sum_c \gamma_c^{(n)} \left( \log \pi_c - \log \gamma_c^{(n)} \right),\\[0.75em]
            \mathcal{L}^{(n)}_{\text{reg}}
                &= \beta\, D_{\mathrm{KL}}\bigl(q_\phi(z \mid x^{(n)}) \,\|\, \mathcal{N}(0,I) \bigr) \\
                &= \frac{\beta}{2} \sum_j \left( \bigl(\mu_j^{(n)}\bigr)^2 + \sigma_j^{2(n)} - 1 - \log \sigma_j^{2(n)} \right),\\[0.75em]
            \mathcal{L}^{[b]}_{\text{ELBO}} &= \sum_{n\in B_b} \mathcal{L}^{(n)}_{\text{ELBO}} , \quad \mathcal{L}^{[b]}_{\text{reg}} = \sum_{n\in B_b} \mathcal{L}^{(n)}_{\text{reg}} ,
        \end{align*}
        where $c$ is the index for GMM clusters, $j$ is the index for latent dimensions, and the additional
        $\beta$–weighted KL term $\mathcal{L}^{(n)}_{\text{reg}}$ regularizes the latent manifold toward a neighborhood of the standard normal prior $\mathcal{N}(0,I)$.

    \State \textbf{Perform ADAM gradient descent} on \( \mathcal{L}^{[b]} = -   \mathcal{L}^{[b]}_{\text{ELBO}} +    \mathcal{L}^{[b]}_{\text{reg}} \)
    \EndFor
        \For{$t = 1$ to $N_{\mathrm{EM}}$}   
            \Comment{Update GMM via EM Step}
                \begin{align*}
                    \gamma_c^{(n)} &\propto \pi_c \, \mathcal{N}(z^{(n)} \mid \mu_c, \sigma_c^2 I), \qquad
                    \gamma_c^{(n)} \gets \frac{\gamma_c^{(n)}}{\sum_{c'} \gamma_{c'}^{(n)}} \\
                    \mu_c &\gets \frac{ \sum_n \gamma_c^{(n)} \, \mu^{(n)} }{ \sum_n \gamma_c^{(n)} }, \\
                    \sigma_c^2 &\gets \frac{ \sum_n \gamma_c^{(n)} \left( (\mu^{(n)} - \mu_c)^2 + \sigma^{2(n)} \right) }{ \sum_n \gamma_c^{(n)} } 
                \end{align*}
        \EndFor
\EndWhile
\end{algorithmic}
\end{algorithm}

\subsection{Model Architecture}
\label{subsec:model_architecture}

We adopt a convolutional encoder and decoder architecture inspired by the U-Net, with modifications for processing Navier-Stokes flow fields and Schlieren images. Implementation details can be found in~\ref{sec:impl-details}. In our proof-of-concept, we employ a Multilayer Perceptron (MLP)-based encoder and decoder.

\section{Evaluation Metric for Interpretability}
\label{sec:evaluation_metric}


To achieve \emph{physical interpretability} in learned latent representations, we define interpretability as the requirement that continuous physical quantities—such as pressure, temperature, or species concentration—vary smoothly with respect to the latent coordinates. Under this definition, small movements in the latent space should correspond to small, physically meaningful changes in measured quantities. This provides a concrete and quantitative criterion for evaluating whether a dimension reduction method preserves the underlying physical continuity of the system.

In this section, we introduce a metric that quantifies this form of interpretability by assessing the smoothness of physical variables across the latent manifold using tools from graph spectral theory.

\subsection{Preliminary: Graph Spectral Theory}

Spectral graph theory provides standard tools for quantifying the smoothness of functions defined on graphs~\cite{chung1997spectral,belkin2003laplacian}.
The Laplacian matrix \( L = D - A \) is defined in terms of the adjacency matrix \(A\) and the degree matrix \(D\), where \(D\) is a diagonal matrix with entries \(D_{ii} = \sum_j A_{ij}\), i.e., each diagonal element records the (unweighted) number of neighbors of node \(i\) in the \(k\)-nearest-neighbor graph.

The eigenvectors of \( L \) represent orthogonal modes of variation on the graph, with eigenvalues indicating their smoothness. Modes associated with small eigenvalues vary slowly across the graph and correspond to smooth, large-scale structure; those with larger eigenvalues represent high-frequency, rapidly varying patterns.

To connect this framework to interpretability, a physical quantity (e.g., pressure) is evaluated at each latent point over the entire latent manifold, which consists of all experimental trajectories, and treated as a graph signal. Decomposing this signal into the Laplacian eigenbasis reveals how much of its variation lies in smooth versus oscillatory modes. If a latent representation is physically interpretable under our definition, then physically continuous quantities should project predominantly onto low-frequency eigenvectors.

By analyzing graph-spectral smoothness directly on the latent embeddings—rather than reconstructed images—we assess how well the learned latent manifold itself captures the intrinsic physical continuity of the underlying system.

Graph spectral methods can be defined using several Laplacian operators, most commonly the unnormalized Laplacian \(L = D - W\) and the normalized Laplacian \(L_{\mathrm{norm}} = D^{-1/2} L D^{-1/2}\)~\cite{chung1997spectral,shuman2013emerging}.  
Both induce orthonormal eigenbases that serve as graph Fourier modes, where smaller eigenvalues correspond to smoother variations over the graph and larger eigenvalues capture increasingly oscillatory behavior.

The two operators differ in structural and spectral properties.  
The normalized Laplacian rescales edge contributions by vertex degrees, yielding spectra bounded in \([0,2]\) and admitting additional features such as spectral folding on bipartite graphs.  However, its eigenvector associated with the zero eigenvalue need not be constant.  
In contrast, the unnormalized Laplacian always has the constant vector as its zero-eigenvalue eigenfunction in connected graphs, which offers a clear and intuitive baseline for smoothness. 

In this work, we adopt the unnormalized Laplacian to retain this direct low-frequency interpretation. Nevertheless, we note that the normalized Laplacian is a popular choice in analyzing graph-based data and may be preferable in applications where degree normalization is desired.

Applications of graph spectral theory span numerous domains, including signal processing~\cite{shuman2013emerging, Ortega2022}, dimensionality reduction~\cite{belkin2003laplacian}, and machine learning~\cite{bronstein2017geometric}. Leveraging this theory, we can evaluate the smoothness of physical variables across data-driven latent manifolds.

\subsection{Evaluation Metric Definition}

The goal of this evaluation is to quantify the physical interpretability of the learned latent manifold—specifically, how smoothly a physical quantity, such as pressure or temperature, varies across the latent space. A smoother distribution indicates that the latent coordinates capture the underlying physical relationships more faithfully. The evaluation process involves the following steps:

\begin{enumerate}
    \item \textbf{Graph Construction}:  Construct a \( k \)-nearest neighbors (kNN) graph with the latent embeddings as nodes to capture the topology of the latent manifold.
    \item \textbf{Laplacian Computation}: Compute the Laplacian matrix \( L \) of the graph and perform an eigen-decomposition:
    \[
    L v_i = \lambda_i v_i.
    \]
    Here, the eigenvectors \( v_i \) represent modes of variation within the latent space.
    
    \item \textbf{Projection of Physical Quantities}: 
    Project the physical quantity of interest (e.g., pressure) \( p \) onto the eigenvectors to obtain coefficients \( \alpha_i \):
    \[
    \alpha_i = v_i^\top p.
    \]
    This expresses the physical quantity in the spectral domain, reflecting the manifold’s intrinsic geometry.
    
    \item \textbf{Energy Concentration Evaluation}: 
    Let \( \mathcal{I}_r \) denote the index set corresponding to the lowest \(r\%\) of eigenvalues (i.e., the smoothest spectral modes). 
    Compute the fraction of the total energy contained within these modes:
    \[
    \eta = \frac{\sum_{i \in \mathcal{I}_r} \alpha_i^2}{\sum_{i} \alpha_i^2}.
    \]
    A higher \( \eta \) value indicates that the physical distribution is more concentrated and smoothly represented by a limited number of spectral modes, suggesting that the dimension reduction method effectively preserves global physical structures.
\end{enumerate}

This metric provides a means to evaluate how well different dimension reduction techniques maintain the underlying physical properties of the data, offering a benchmark for comparing their effectiveness in preserving physical interpretability.

\section{Experiments and Discussions}


In this section, we validate the GM-VAE on both simulated and experimental datasets. 
We assess the method along three main axes: 
(i) reconstruction accuracy, measured via mean-squared error against reference simulation or experimental data; 
(ii) recovery of physically meaningful regimes, quantified by comparison of latent clusters with known labels (e.g., ignition type, wake regime); 
and (iii) physical interpretability of the latent manifold, evaluated using the graph-spectral smoothness metric defined in Section~\ref{sec:evaluation_metric}. 
For the low-dimensional surface-reaction example, we additionally illustrate the generative capabilities of the GM-VAE by comparing synthetic trajectories with reference solutions. 

The experiments are conducted on:
\begin{itemize}
    \item time-series density data from a surface reaction model with bifurcation outcomes,
    \item velocity fields from flow past two side-by-side square cylinders, and
    \item Schlieren images from laser-induced ignition experiments in a turbulent combustion chamber.
\end{itemize}

\subsection{Surface Reaction}
\label{subsec:experiments_surface}

\paragraph{Dataset}  
We begin with a proof-of-concept study using simulated data generated from a surface reaction model exhibiting bifurcation dynamics~\cite{makeev2002coarse,hampton2015compressive}. This model describes the evolution of a reaction variable \(\rho(t)\), which represents the fractional surface coverage in a reaction-diffusion system, governed by the differential equation
\begin{equation}
\frac{d\rho}{dt} = \alpha(1 - \rho) - \gamma \rho - \kappa \rho(1 - \rho)^2, \quad \rho(0) = 0.89,
\end{equation}
where \(\alpha\), \(\gamma\), and \(\kappa\) control growth, decay, and nonlinear interaction terms, respectively. Here, \(\kappa=1.0\) is constant, while the other parameters are modeled as random variables to introduce variability
\begin{align}
\alpha(\xi_1) &= 0.1 + e^{0.05 \xi_1}, \quad \xi_1 \sim \mathcal{N}(0, 1), \\
\gamma(\xi_2) &= 0.001 + 0.01e^{0.05 \xi_2}, \quad \xi_2 \sim \mathcal{N}(0, 1).
\end{align}

The model produces two outcomes based on the initial conditions and parameter variations:
\begin{itemize}
    \item \textbf{Stable State}: \(\rho(t)\) remains near its initial value or stabilizes at a low equilibrium, indicating minimal reactivity.
    \item \textbf{Reactive State}: \(\rho(t)\) evolves dynamically, often approaching a high equilibrium (e.g., \(\rho \to 1\)), signifying active reaction dynamics.
\end{itemize}

\paragraph{Experiments}  
We simulated 1280 trajectories, each with 50 time steps, representing the temporal evolution of \(\rho(t)\). The dataset was split randomly into 80\% training, 10\% validation, and 10\% testing sets. A GM-VAE model with 2 latent dimensions and 2 Gaussian clusters was then fitted to the training set.

\begin{figure}[htbp]
    \centering
    \includegraphics[width=0.32\linewidth]{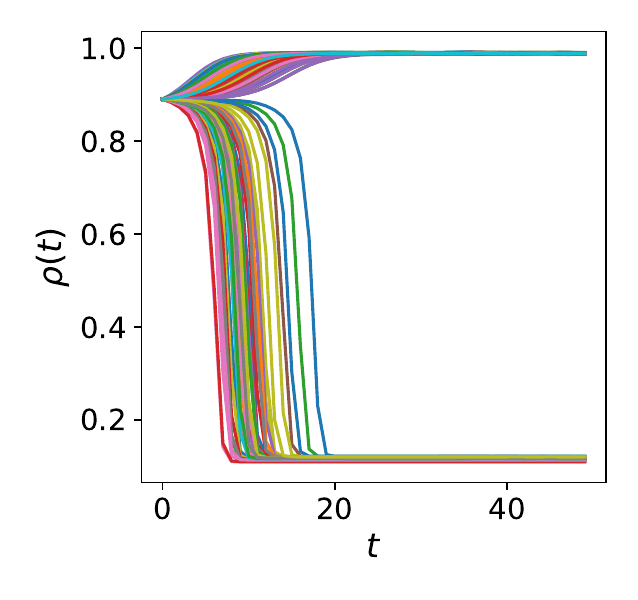}~
    \includegraphics[width=0.32\linewidth]{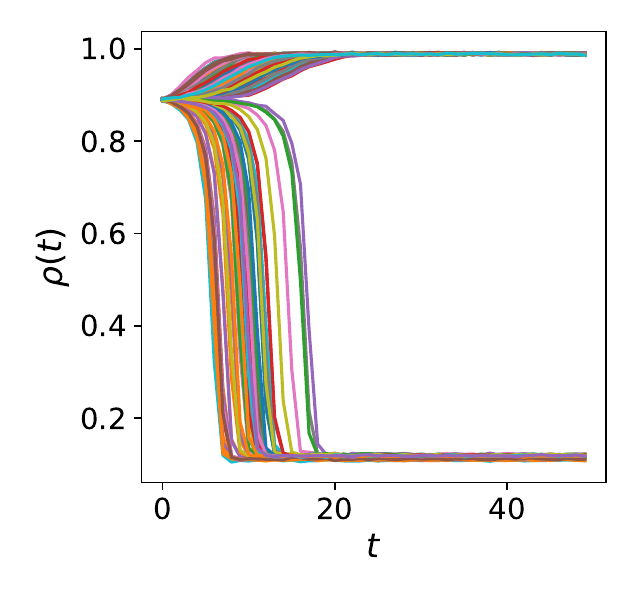}~
    \includegraphics[width=0.323\linewidth]{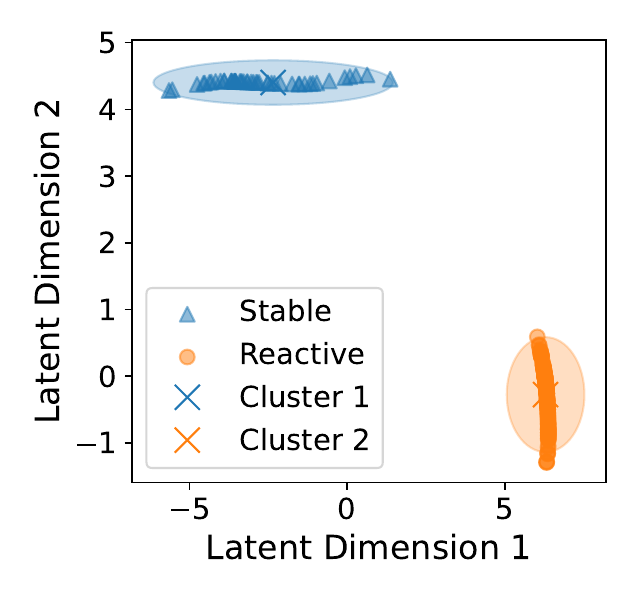}
    \caption{Training samples (left), generated samples (middle), and latent representations (right) for surface reaction simulations. The GM-VAE compresses the data into 2 latent dimensions and achieves perfect clustering of the bifurcation outcomes.}
    \label{fig:surface_reaction_results}
\end{figure}

\paragraph{Results}  
The GM-VAE successfully compressed the high-dimensional trajectories into a 2D latent space, achieving perfect clustering accuracy for the bifurcation outcomes. \Cref{fig:surface_reaction_results} illustrates:
\begin{itemize}
    \item \textbf{Training Samples}: Time-series curves of \(\rho(t)\) for both stable and reactive states (left).
    \item \textbf{Generated Samples}: High-quality synthetic trajectories generated from the latent space (middle) that closely resemble the training data.
    \item \textbf{Latent Representations}: Clear separation between clusters representing stable and reactive states (right), highlighting the model's ability to capture the underlying bifurcation dynamics.
\end{itemize}

\begin{figure}[htbp]
    \centering
    \includegraphics[width=0.6\linewidth]{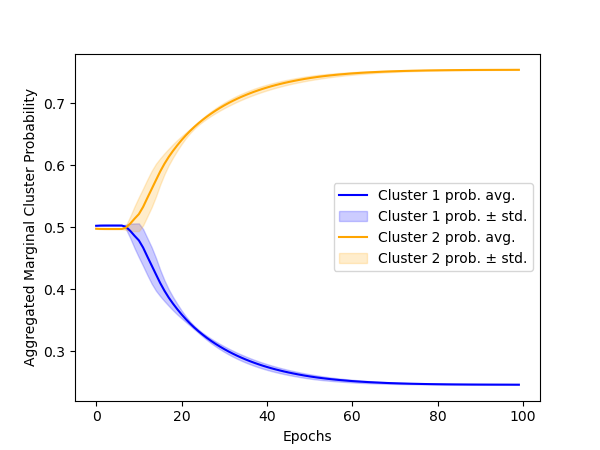}
    \caption{The evolution of the neural network predicted marginal cluster probability $\pi$, aggregated across trainings with different random initializations.}
    \label{fig:surface_reaction_cluster_prob_convergence}
\end{figure}

Moreover, the GMM assumption facilitates both conditional and unconditional sampling. Given a class label, the model enables conditional sampling by decoding a sample from the corresponding Gaussian cluster. Its perfect clustering accuracy provides an accurate estimation of the cluster probabilities, allowing unconditional sampling by first drawing a class label and then a corresponding data point.

\subsection{Flow Past Square Cylinders}
\label{subsec:experiments_cylinder}

\begin{figure}[htbp]
    \centering
    \includegraphics[width=1\linewidth]{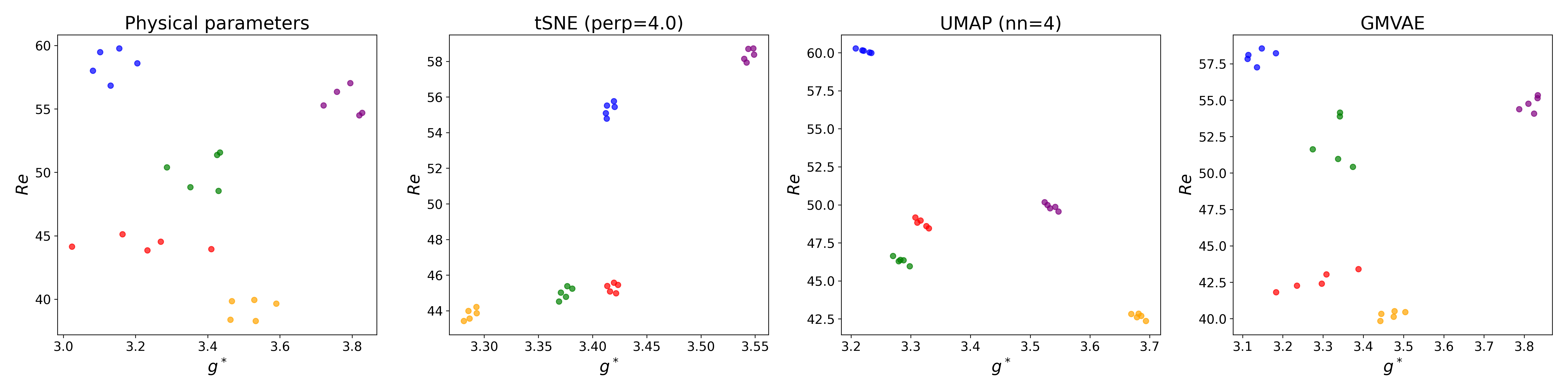}
    \caption{Physical parameters of the simulations and 2D latent embeddings learned by tSNE, UMAP, and GM-VAE (up to an affine transformation computed from solving a linear regression problem, see details in Appendix~\ref{sec:affine-mapping}). GM-VAE recovers the physical parameter space from data-driven learning by learning a latent manifold where the axes closely correlate with key physical quantities.}
    \label{fig:two-square-cylinders-label}
\end{figure}

\paragraph{Dataset}
We employ velocity field data obtained from numerical simulations of the wake flow behind two identical side-by-side square cylinders at low Reynolds numbers, as detailed in the problem setup presented in Figure 1 of~\cite{ma2017wake}. The numerical simulation process is described in~\cite{lee2025surrogate}. The computational domain is defined relative to the cylinder dimension 
$D$; two identical square cylinders are placed symmetrically in the 
$y$-direction with their gap spacing 
$g$ varying such that the nondimensional gap ratio 
$g^{*}=g/D$ spans between 2.975 and 4.025. Each snapshot includes multichannel flow fields consisting of $u$ (x-velocity) and
$v$ (y-velocity). The dataset covers a range of laminar wake regimes corresponding to Reynolds numbers up to 70. The physical system is controlled by two parameters, the size of the gap, denoted $g^*$, and the Reynolds number, denoted $Re$. The velocity fields exhibit qualitatively different characteristics according to different combinations of $(g^*, Re)$ values. We generated discrete labels corresponding to the characteristic regions and randomly generated 5 $(g^*, Re)$ sample pairs from 5 distinct regions. The resulting dataset is plotted in the left panel, and labels are indicated by colors in \Cref{fig:two-square-cylinders-label}.

\paragraph{Experiments}
A GM-VAE with a convolutional encoder and decoder was trained on this dataset using a 2-dimensional latent space. This architecture was chosen to efficiently capture the subtle variations in wake dynamics as functions of both Reynolds number and gap ratio.

\begin{figure}[htbp]
    \centering
    \includegraphics[width=1\linewidth]{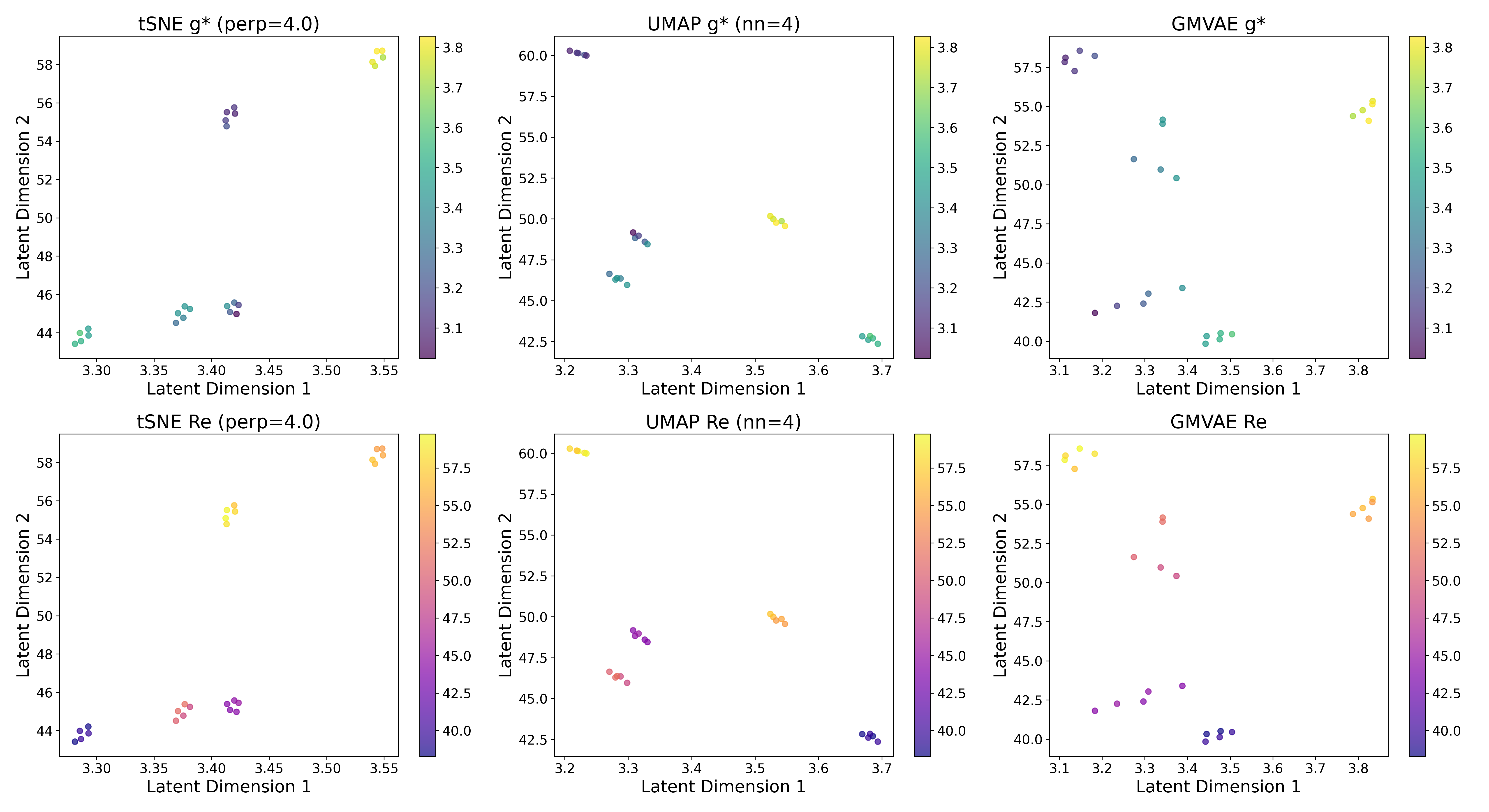}
    \caption{Gap size and Reynolds number visualized in the 2D latent manifold learned by t-SNE, UMAP, and GM-VAE. }
    \label{fig:two-square-cylinders-params}
\end{figure}

\paragraph{Results}
Figure~\ref{fig:two-square-cylinders-label} illustrates that the GM-VAE latent manifold recovers the data distribution with respect to gap ratio and Reynolds number in the space of physically meaningful parameters. The GMM cluster centers align with characteristic wake patterns. The tSNE method is designed with a substantial focus on local neighborhood relationships, and thus tends to distort global relationships in favor of local structure. It is also less robust with respect to cluster size. UMAP constructs a low‐dimensional embedding by approximating the high‐dimensional data manifold with a fuzzy simplicial complex, then optimizes a cross‐entropy objective to preserve both local and global structure. For the laminar flow dataset, UMAP shows sensitivity to the choice of the number of neighbors. Although UMAP often separated major flow regimes better than tSNE, its cluster centroids did not align as precisely with the data modes as the Gaussian mixture model centroids learned by GM-VAE. Compared to tSNE and UMAP, GM-VAE finds a more physically meaningful latent space from data-driven learning by balancing both local and global structure of the data. GM-VAE also learns a latent manifold where physical parameters are smoother in both latent coordinates. 

\subsection{Turbulent Combustion}
\label{subsec:experiments_schlieren}

\begin{figure}[htbp]
    \centering
    \includegraphics[width=1.0\linewidth]{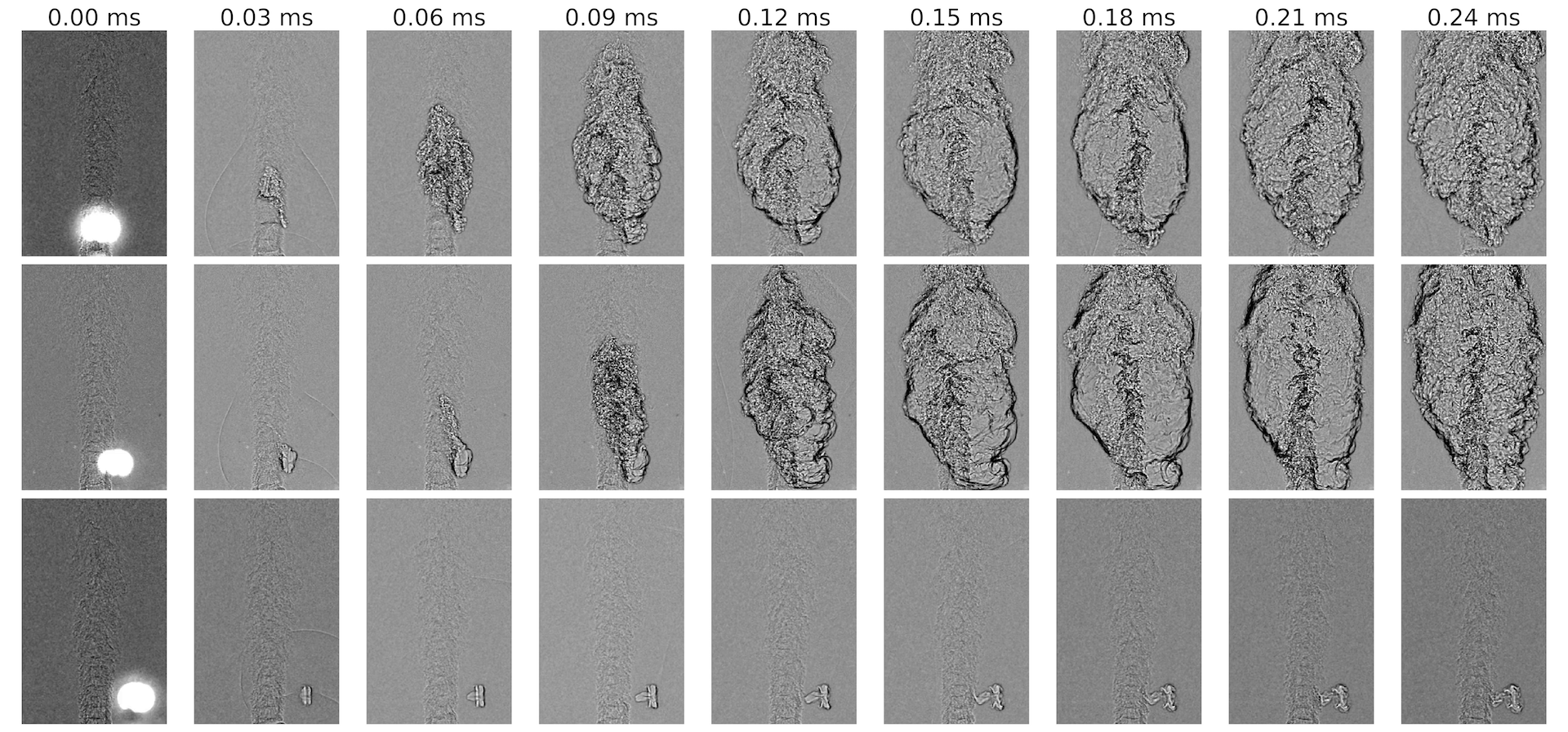} 
    \caption{Schlieren image time-series from a directly ignited experiment (top), an indirectly ignited experiment (middle), and a non-ignited experiment (bottom).}
    \label{fig:schlieren_data_example}
\end{figure}

\paragraph{Dataset}  
We analyze a dataset of Schlieren images from laser-induced ignition experiments. The combustion process is highly stochastic due to complex interactions between combustion chemistry and fluid dynamics. The dataset comprises 2950 Schlieren images from 118 experiments, covering different ignition types: directly ignited, indirectly ignited, and non-ignited. Snapshots are captured at a time step of 0.01 ms. Examples from each ignition type are shown in \Cref{fig:schlieren_data_example}.

\begin{figure}[htbp]
    \centering
    \includegraphics[width=1.0\linewidth]{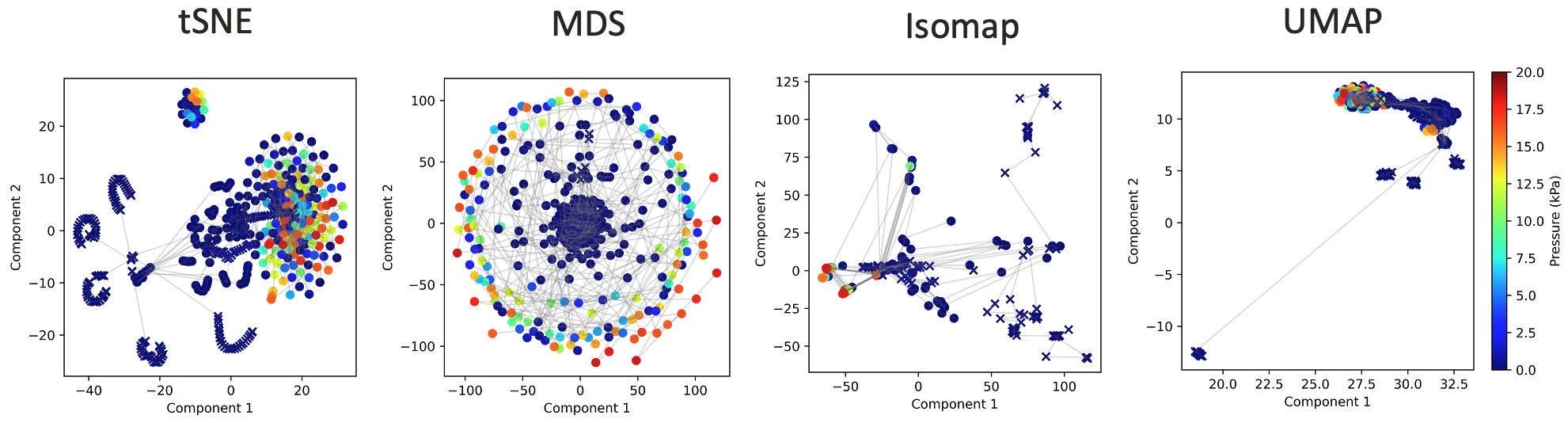}
    \caption{2D representations of Schlieren images produced by classical nonlinear dimension reduction methods.}
    \label{fig:classical_nonlinear}
\end{figure}

\paragraph{Baseline Representations}  
To establish a benchmark, we first applied classical nonlinear dimension reduction methods (e.g., t-SNE, MDS, Isomap, UMAP) to the combustion dataset. \Cref{fig:classical_nonlinear} shows the resulting 2D representations. While these methods preserve local pixel-domain similarities, they struggle to separate frames from different experiments and fail to cluster frames based on similar physical states, thereby exposing limitations in capturing meaningful patterns in noisy, high-dimensional data.

\begin{figure}[htbp]
    \centering
    \includegraphics[width=1.0\linewidth]{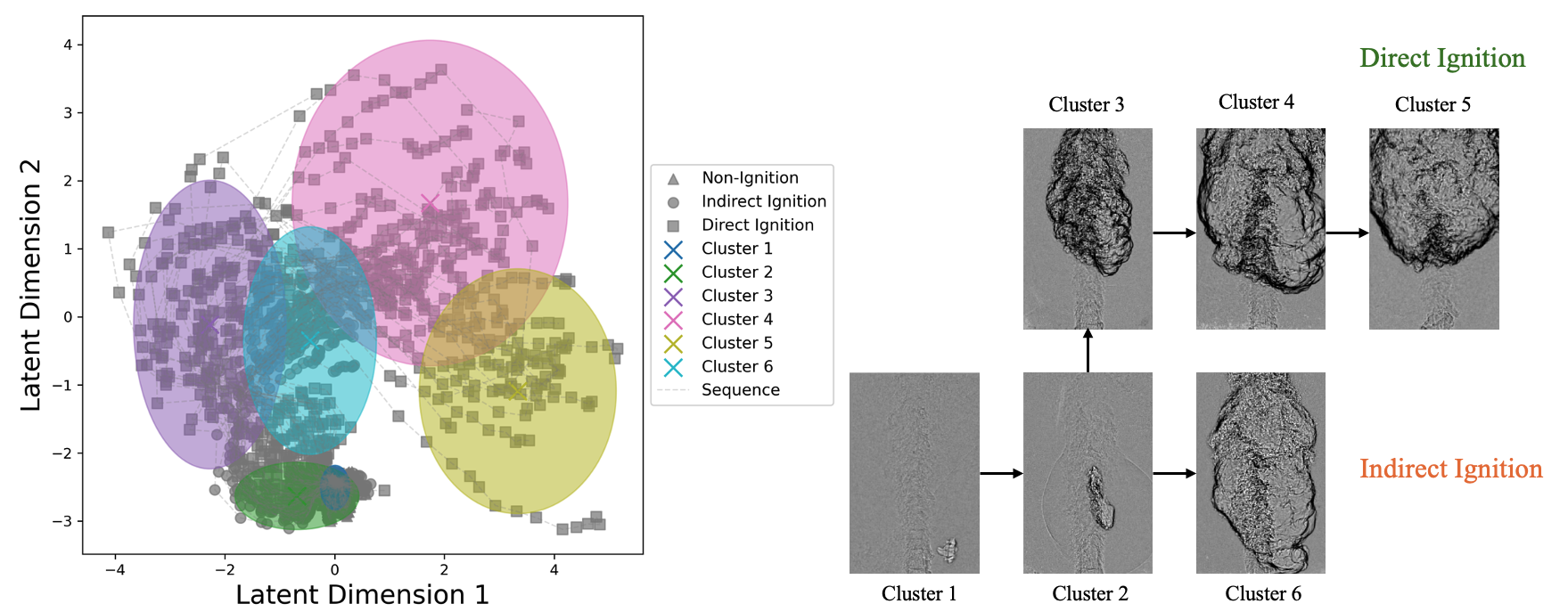} 
    \caption{2D latent representations of Schlieren images generated by the GM-VAE. Non-ignited experiments form dense clusters, while ignited experiments trace distinct trajectories.}
    \label{fig:gmvae_latent}
\end{figure}

\paragraph{Discovery of Ignition Paths}  
We applied a GM-VAE model to the Schlieren image dataset. The resulting 2D latent representations (\Cref{fig:gmvae_latent}) reveal that images from non-ignited experiments form dense clusters, whereas images from ignited experiments trace distinct trajectories across the latent space. The GM-VAE identifies characteristic paths given the time span of the collected experimental data: direct ignition trajectories progress through clusters 1 \(\to\) 2 \(\to\) 3 \(\to\) 4 \(\to\) 5, while indirect ignition trajectories follow a path through clusters 1 \(\to\) 2 \(\to\) 6. These embeddings exhibit smooth variations in physical measurements (e.g., pressure, timestamp, spark location), underscoring the model's ability to capture the underlying physical states despite not having these inputs during training.

\begin{figure}[htbp]
    \centering
    \includegraphics[width=0.39\linewidth]{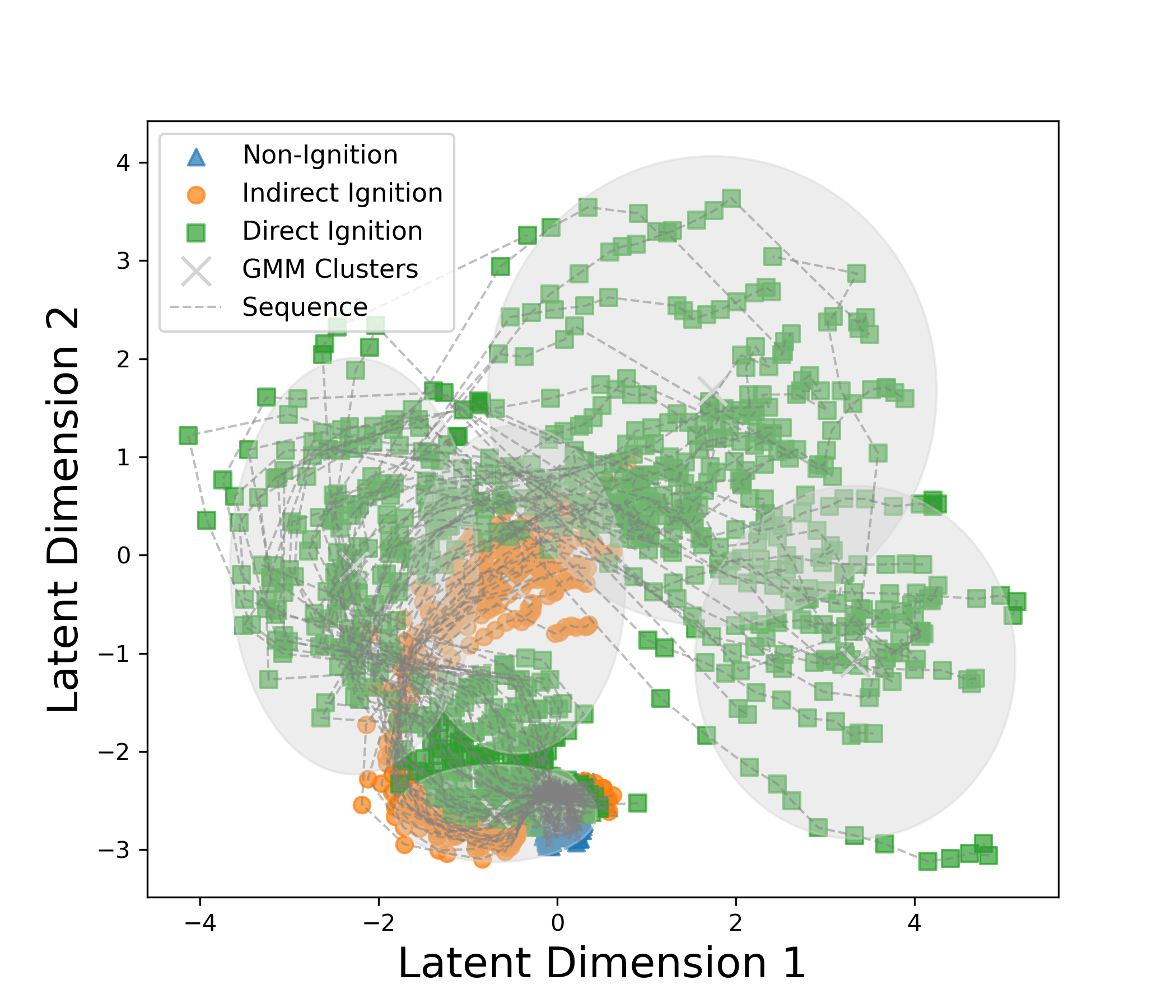}\hspace{0.8cm}
    \includegraphics[width=0.45\linewidth]{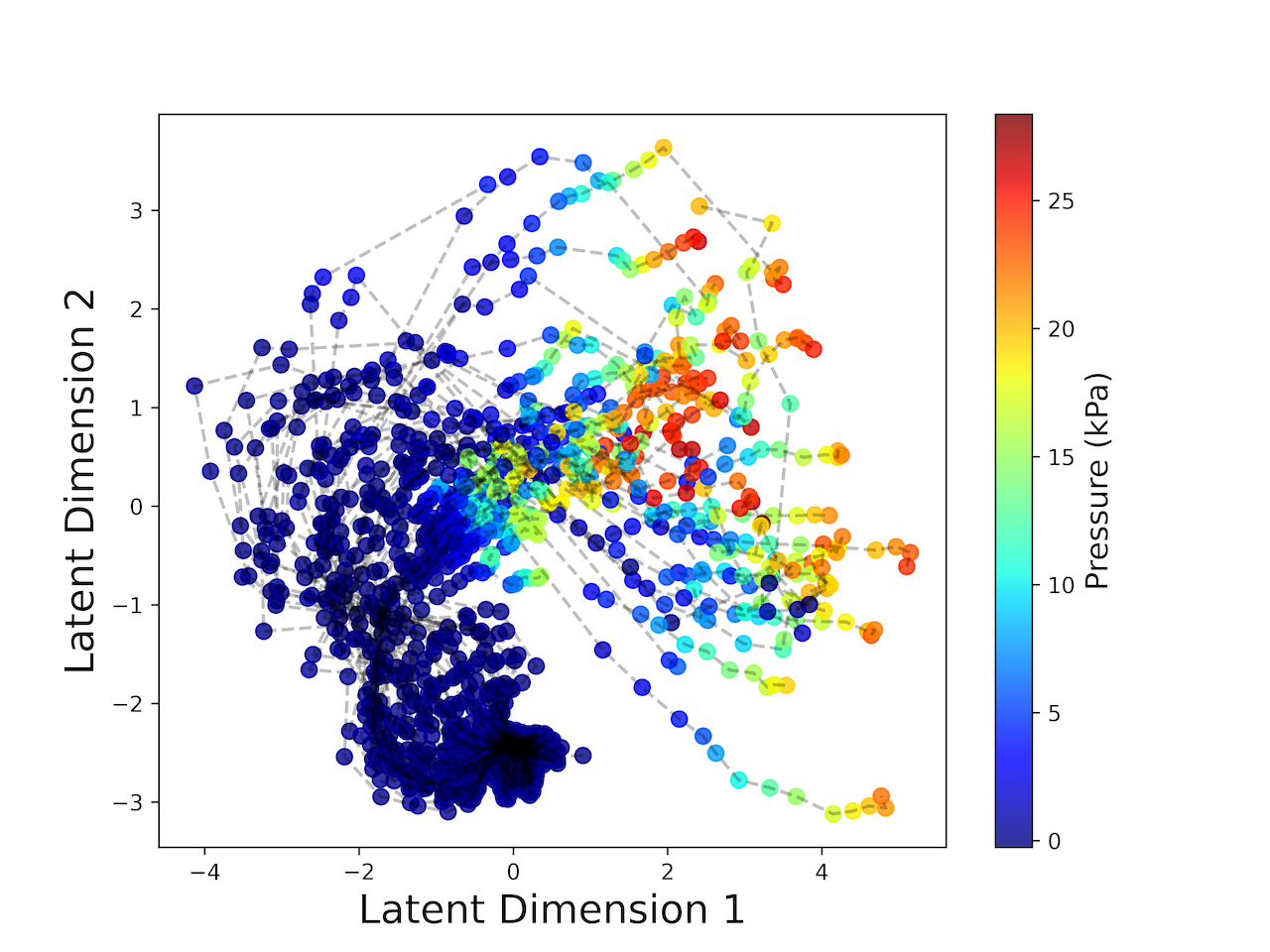}\\[1ex]
    \includegraphics[width=0.45\linewidth]{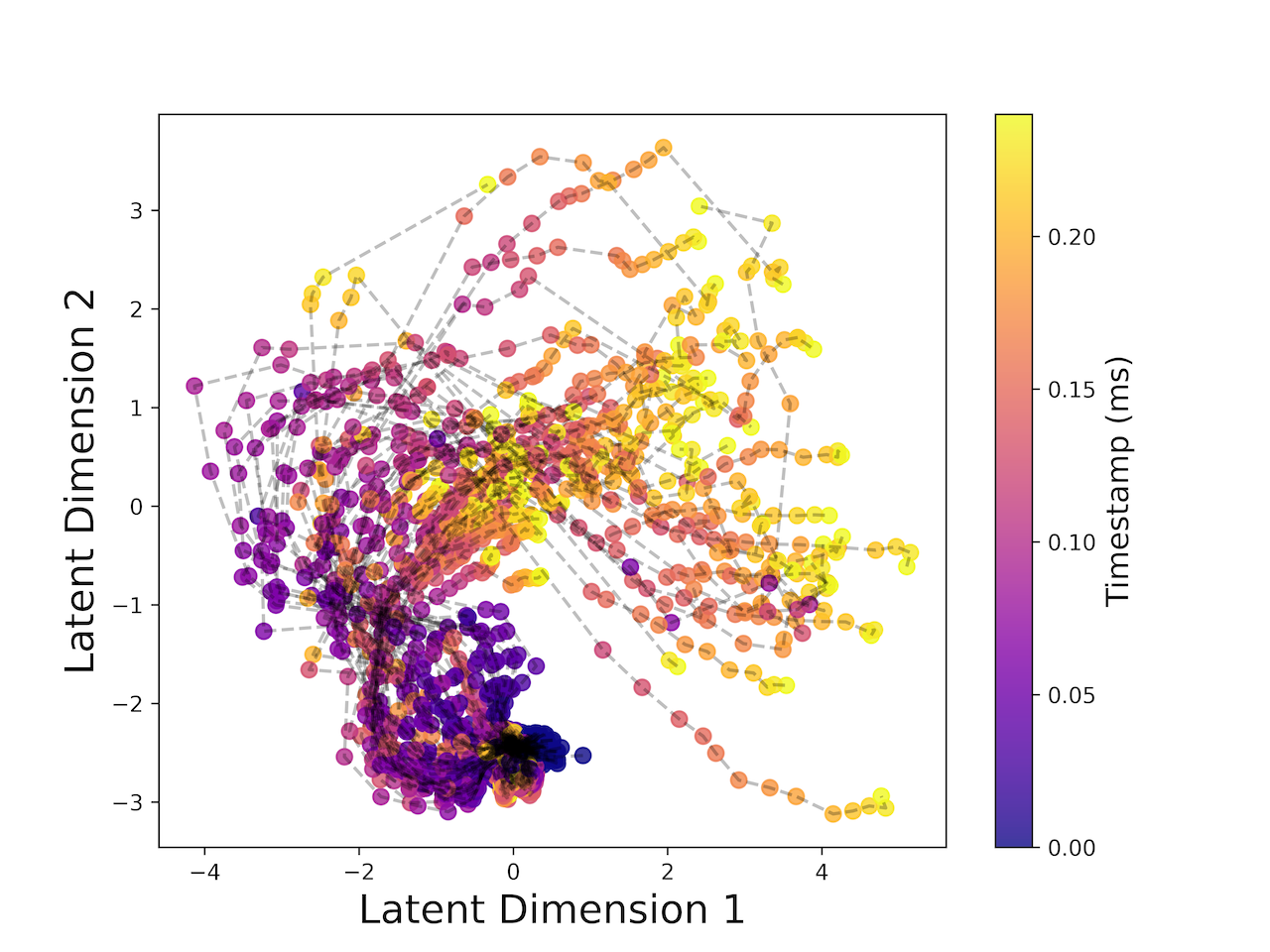}\hspace{0.8cm}
    \includegraphics[width=0.45\linewidth]{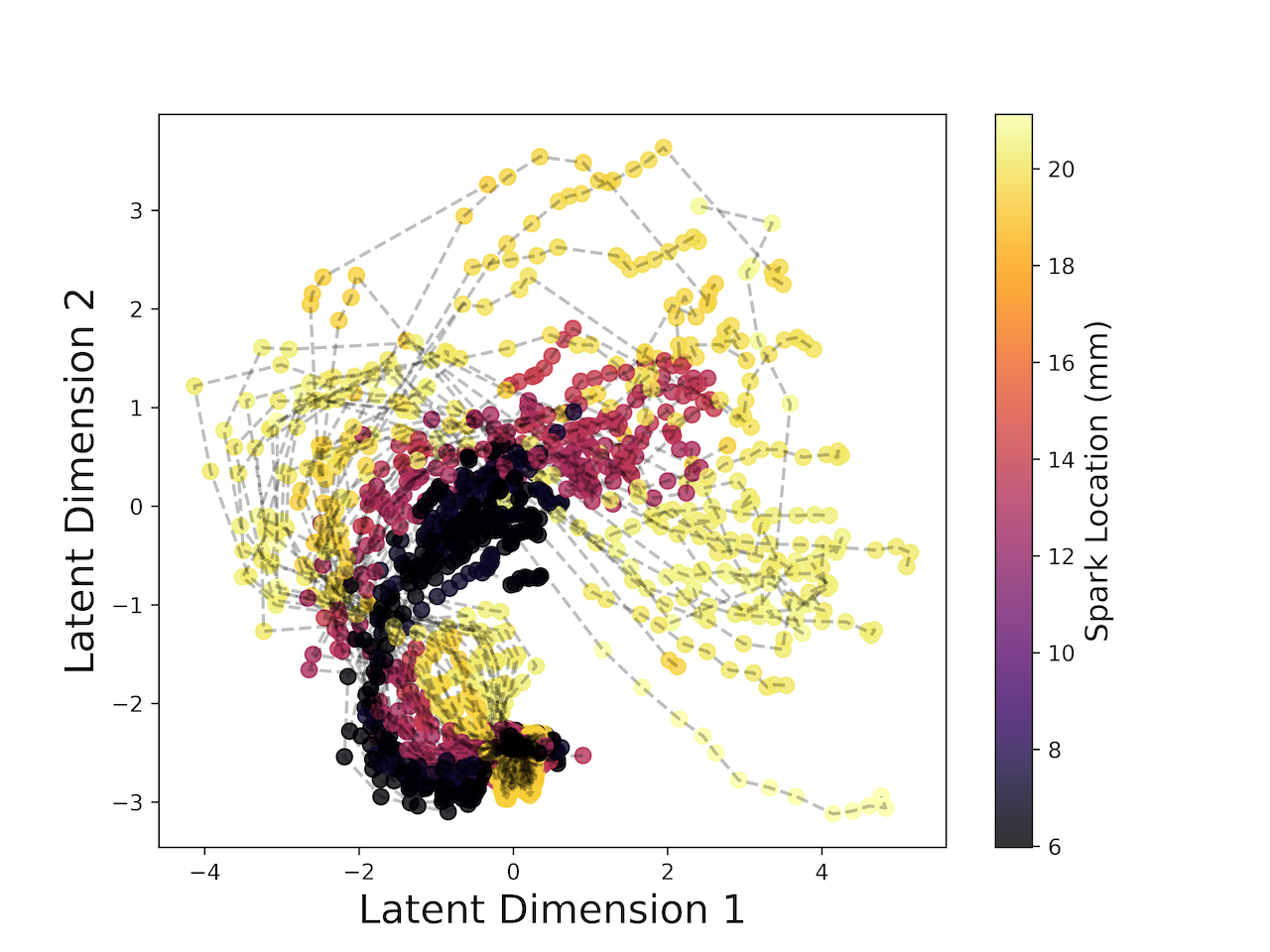}
    \caption{2D GM-VAE embeddings with smooth variations in physical measurements. The subfigures illustrate embeddings colored by ignition types, pressure, timestamp, and spark location.}
    \label{fig:gmvae_by_all}
\end{figure}

\begin{figure}[htbp]
    \centering
    \includegraphics[width=\linewidth]{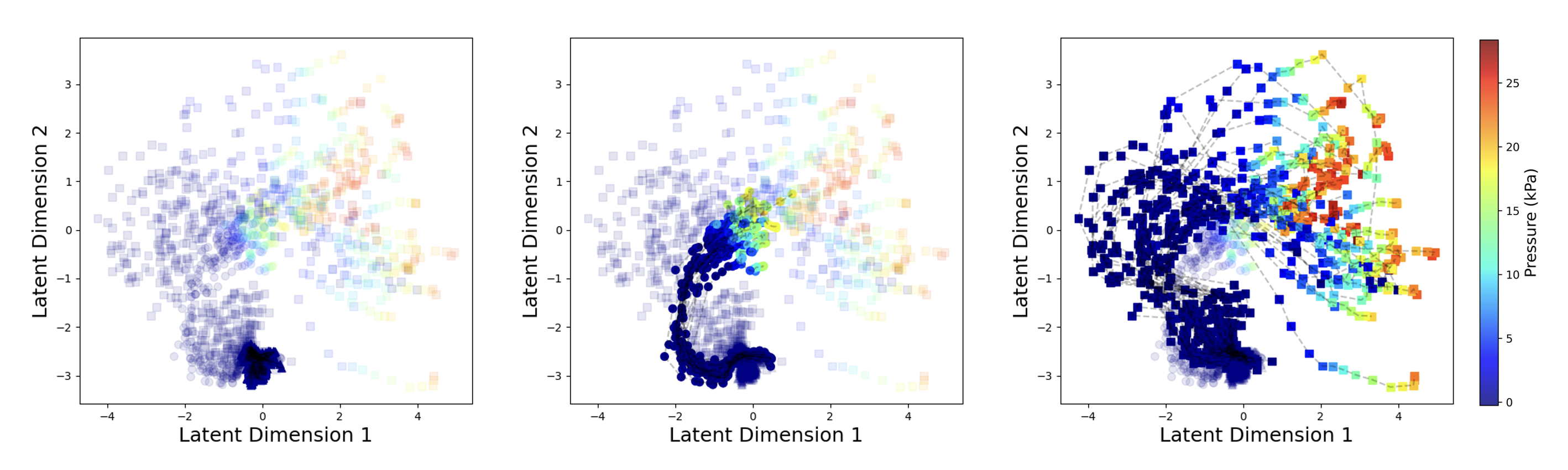} 
    \caption{Visualization of pressure values mapped onto the GM-VAE latent manifold. Embeddings corresponding to non-ignition, indirect ignition, and direct ignition are shown.}
    \label{fig:gmvae_by_ignition}
\end{figure}

\begin{figure}[htbp]
    \centering
    \includegraphics[width=0.65\linewidth]{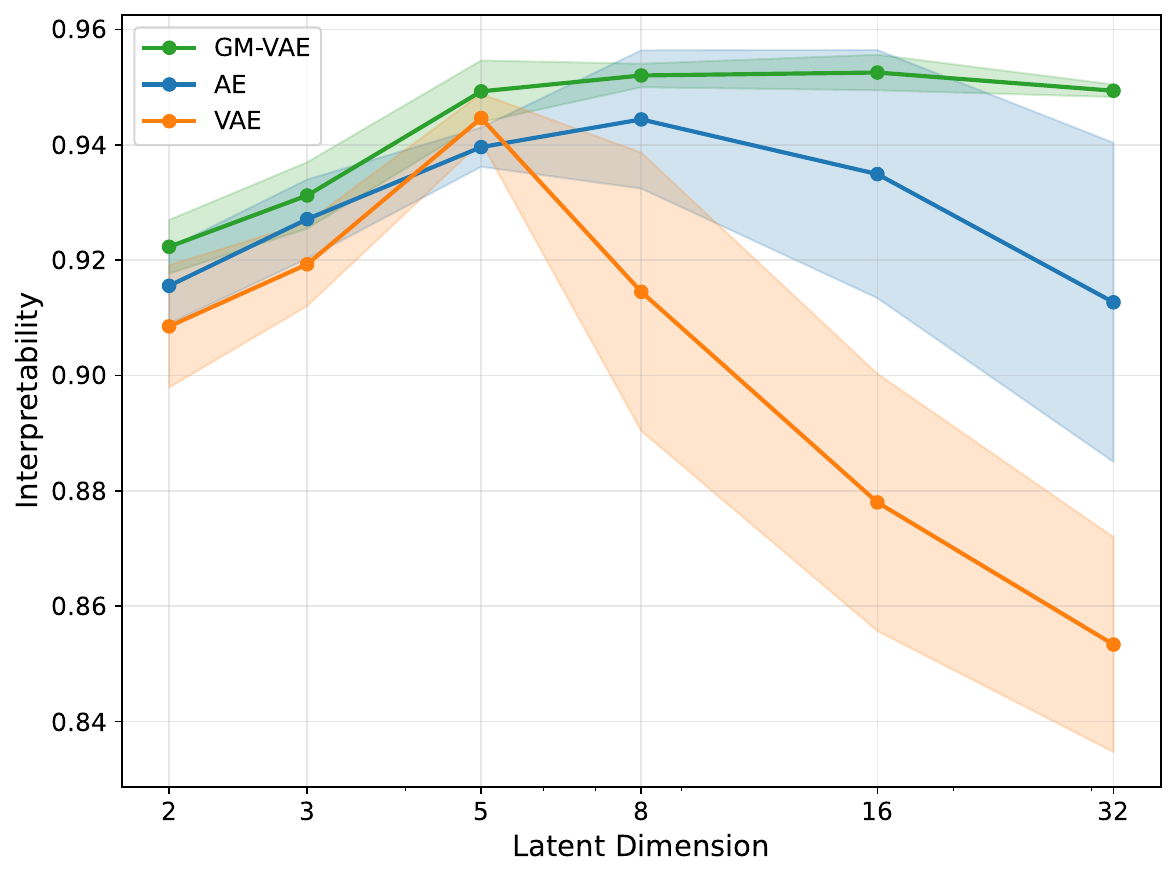}
    \caption{GM-VAE consistently outperforms baseline AE and VAE models in terms of physical interpretability.}
    \label{fig:compare}
\end{figure}

\paragraph{Physical Interpretability}  
As discussed in \Cref{sec:evaluation_metric}, we assess physical interpretability based on the model's ability to uncover a latent manifold where physical quantities (excluded during training) vary smoothly with the latent coordinates. Our results indicate that the GM-VAE consistently yields higher physical interpretability compared to baseline methods. A quantitative comparison is included in Figure \ref{fig:compare}, with $k=100$ for kNN graph construction and $r=20$ top percentage of the Laplacian eigenmodes for the evaluation procedure described in \Cref{sec:evaluation_metric}.

\paragraph{Number of Clusters}  
The GM-VAE embeddings are robust to variations in the number of clusters. With a single cluster, the model reduces to a standard VAE, and embeddings collapse near the origin. As the number of clusters exceeds 3, latent space trajectories emerge, reflecting the sequential nature of the data. Increasing the number of clusters beyond 6 captures finer distinctions among samples, enhancing the representation of subtle variations in physical states. There tends to be a diminishing return in interpretability as the number of clusters increases beyond 9.

\paragraph{Latent Space Dimension}  
As the latent space dimension increases beyond 8, we observe a trade-off between reconstruction accuracy and interpretability. Lower dimensions facilitate the disentanglement of latent embeddings, whereas higher dimensions improve reconstruction fidelity by preserving finer details.

\section{Conclusions}

In this work, we presented a Gaussian Mixture Variational Autoencoder (GM-VAE) framework designed to extract physically interpretable, low-dimensional representations from high-dimensional scientific data. By integrating an EM-inspired block-coordinate descent training strategy, our approach stabilizes the joint optimization of reconstruction and clustering, overcoming the training instabilities common in standard VAEs. We validated the framework across a hierarchy of physical systems---from bifurcating surface reaction ODEs to chaotic wake flows and experimental laser-induced combustion Schlieren images.

Crucially, we introduced a novel spectral interpretability metric based on graph Laplacian smoothness. This quantitative benchmark allowed us to objectively assess whether the learned latent topology respects the continuity of governing physical variables. Our experiments demonstrated that the GM-VAE significantly outperforms baseline methods (t-SNE, UMAP, and standard VAEs) by organizing data into topologically consistent manifolds where latent clusters align with distinct physical regimes---such as stable vs.\ unstable vortex shedding or direct vs.\ indirect ignition pathways.

Despite its strengths, the framework presents limitations. First, the computational complexity is higher than that of standard VAEs due to the iterative EM steps. Second, the clustering performance remains sensitive to initialization, requiring a brief pre-training phase to ensure convergence. Finally, the assumption that physical regimes map to Gaussian distributions in the latent space may be restrictive in some cases.

Looking forward, the ability to generate physically grounded latent representations opens new avenues for scientific discovery. In fluid dynamics, this framework offers a data-driven method for identifying flow regimes without manual labeling; in combustion, it enables the unsupervised classification of stochastic ignition events, facilitating better diagnostic tools. Our future work will extend the GM-VAE to multi-modal learning, incorporating diverse data streams such as pressure gradients, chemical concentrations, and laser positions. By leveraging separate encoders for each modality and fusing representations via the Product of Experts method~\cite{trask2022unsupervised}, we aim to disentangle latent factors that capture cross-modal correlations, further bridging the gap between deep generative modeling and physical interpretability.

\section*{Acknowledgments}

The authors are grateful to Prof.\ Calson Slabaugh's group at Purdue University for conducting the combustion experiments and recording Schlieren imaging data. The authors are grateful to Jiyoung Lee at the University of Melbourne for generously sharing the simulation data of the flow-past-two-cylinders benchmark. The authors also extend their gratitude to Dr.\ Diego Rossinelli, Prof.\ Gianluca Iaccarino, Dr.\ Tony Zahtila, Dr.\ Charlelie Laurent, Dr.\ Davy Brouzet, Dr. \ Donatella Passiatore, and Dr.\ Juan Cardenas for helpful discussions over the course of this work.

This material is based upon work supported by the Department of Energy's National Nuclear Security Administration under Award Number DE-NA0003968.



\newpage
\appendix

    

\section{Derivation of the Loss Function}
\label{sec:elbo-derivation}

Under the model assumptions, the per-example ELBO takes the closed form:
\begin{align*}
    & \quad \  \ell_{\text{ELBO}}(x) = \mathbb{E}_{q}\big[ \log  p(x, z, c ) \big] - \mathbb{E}_{q}\big[ \log q(z, c |x ) \big]   \\
    & = \mathbb{E}_{q}\big[ \log  p(x |z ) \big] + \mathbb{E}_{q}\big[ \log  p(z |c ) \big] + \mathbb{E}_{q^{(c)}}\big[ \log  p(c ) \big] - \mathbb{E}_{q^{(z)}}\big[ \log q^{(z)}(z |x ) \big] \\ & \quad - \mathbb{E}_{q^{(c)}}\big[ \log q^{(c)}(c |x ) \big]  \\
    & =  \mathbb{E}_{q^{(z)}}\big[ \log  p(x |z ) \big] + \mathbb{E}_{q}\big[ \log  p(z |c ) \big] + \mathbb{E}_{q^{(c)}}\big[ \log  p(c ) \big] - \mathbb{E}_{q^{(z)}}\big[ \log q^{(z)}(z |x ) \big]  \\ & \quad - \mathbb{E}_{q^{(c)}}\big[ \log q^{(c)}(c |x ) \big] \\
    & = \int_z q^{(z)}(z|x) \log p(x |z ) dz + \sum_c q^{(c)}(c|x) \int_z q^{(z)}(z|x) \log p(z|c) dz  \\ & \quad + \sum_c q^{(c)}(c|x) \log p(c) - \int_z q^{(z)}(z|x) \log q^{(z)}(z|x)  dz - \sum_c q^{(c)}(c|x) \log q^{(c)}(c|x) \\
    & = \int_z q^{(z)}(z|x) \log p(x |z ) dz   -\frac{1}{2}  \sum_c \gamma_c \sum_{j} \Big( \log 2 \pi \sigma_{c}^{2}|_j+\frac{\sigma^{2}|_j}{\sigma_{c}^{2}|_j}+\frac{\left(\mu|_j-\mu_{c}|_j\right)^{2}}{\sigma_{c}^{2}|_j} \Big) \\
    & \quad  
    +\frac{1}{2} \sum_{j} \big( \log 2 \pi \sigma^2 |_j+1 \big)+ \sum_c \gamma_c  \big( \log \pi_c  - \log \gamma_c \big) \\
    & =  \log p(x |z )   -\frac{1}{2}  \sum_c \gamma_c \sum_{j} \Big( \log 2 \pi \sigma_{c}^{2}|_j+\frac{\sigma^{2}|_j}{\sigma_{c}^{2}|_j}+\frac{\left(\mu|_j-\mu_{c}|_j\right)^{2}}{\sigma_{c}^{2}|_j} \Big) \\
    & \quad  
    +\frac{1}{2} \sum_{j} \big( \log 2 \pi \sigma^2 |_j+1 \big)+ \sum_c \gamma_c  \big( \log \pi_c  - \log \gamma_c \big)  
\end{align*} where $\cdot|_j$ denotes the $j$-th entry of a vector.

\section{Classical Non-linear Dimension Reduction Methods}
\label{sec:classical-nonlinear}

\subsection{Multidimensional Scaling (MDS)}
MDS aims to preserve pairwise distances between points in the high-dimensional space in a low-dimensional embedding. Given a high-dimensional pairwise distance matrix \( D_{ij} = \| \mathbf{x}_i - \mathbf{x}_j \|_2 \), the objective is to find a low-dimensional embedding \( \mathbf{y}_i \in \mathbb{R}^d \) (where \( d < D \)) by minimizing the stress function:

\[
\text{Stress} = \sum_{i < j} \left( D_{ij} - \| \mathbf{y}_i - \mathbf{y}_j \|_2 \right)^2.
\]

This optimization can be solved via eigenvalue decomposition or iterative methods.

\subsection{Isomap}
Isomap extends MDS by preserving geodesic distances rather than Euclidean distances, making it suitable for nonlinear manifolds. The process involves constructing a neighborhood graph, computing the shortest path (geodesic) distance \( G_{ij} \) between points, and then applying MDS on the geodesic distance matrix:

\[
\text{Stress} = \sum_{i < j} \left( G_{ij} - \| \mathbf{y}_i - \mathbf{y}_j \|_2 \right)^2.
\]

\subsection{t-Distributed Stochastic Neighbor Embedding (t-SNE)}
t-SNE aims to preserve local structures by minimizing the divergence between pairwise similarities in high- and low-dimensional spaces. The pairwise similarities in the high-dimensional space are modeled by the conditional probability:

\[
P_{ij} = \frac{\exp\left(-\frac{\| \mathbf{x}_i - \mathbf{x}_j \|^2}{2\sigma_i^2}\right)}{\sum_{k \neq l} \exp\left(-\frac{\| \mathbf{x}_k - \mathbf{x}_l \|^2}{2\sigma_k^2}\right)},
\]

while in the low-dimensional space, they are modeled using:

\[
Q_{ij} = \frac{\left( 1 + \| \mathbf{y}_i - \mathbf{y}_j \|^2 \right)^{-1}}{\sum_{k \neq l} \left( 1 + \| \mathbf{y}_k - \mathbf{y}_l \|^2 \right)^{-1}}.
\]

The goal is to minimize the Kullback-Leibler divergence:

\[
\text{KL}(P \| Q) = \sum_{i \neq j} P_{ij} \log \frac{P_{ij}}{Q_{ij}},
\]

using gradient descent to find the low-dimensional embedding \( \mathbf{y}_i \).

\subsection{Uniform Manifold Approximation and Projection (UMAP)}
UMAP balances local and global structure preservation by approximating the manifold structure in the low-dimensional space. It constructs a local fuzzy simplicial set representing the probability that points \( i \) and \( j \) are connected in the manifold. The local weight for the connection between points \( i \) and \( j \) is defined as:

\[
w_{ij} = \exp\left(- \frac{\| \mathbf{x}_i - \mathbf{x}_j \|^2}{\sigma_i^2}\right),
\]

where \( \sigma_i \) is a scaling parameter controlling local connectivity. These local weights \( w_{ij} \) are combined to form a global fuzzy topological structure.

The low-dimensional embedding \( \mathbf{y}_i \) is obtained by minimizing a cross-entropy-like loss function that compares the high-dimensional local weights to their counterparts in the low-dimensional space:

\[
\text{Loss} = \sum_{(i,j)} \left( - w_{ij} \log \sigma(\| \mathbf{y}_i - \mathbf{y}_j \|) - (1 - w_{ij}) \log\left(1 - \sigma(\| \mathbf{y}_i - \mathbf{y}_j \|)\right) \right),
\]

where \( \sigma(\cdot) \) is the logistic function. Optimization is typically carried out using stochastic gradient descent to find the embedding \( \mathbf{y}_i \) that best preserves both local and global structures of the original high-dimensional data.

\section{Mapping Learned Embeddings to Physical Parameters}
\label{sec:affine-mapping}

To facilitate quantitative comparison between different dimensionality reduction methods (e.g., t-SNE, UMAP, and GM-VAE) and to assess how well learned latent representations align with known physical parameters, we employ a post-hoc affine transformation that maps embeddings to the physical parameter space. This approach enables direct visualization of how well each method recovers the underlying physical structure, as measured by the correspondence between latent coordinates and physical quantities such as gap ratio $g^*$ and Reynolds number $Re$.

\subsection{Mathematical Formulation}

Given a set of $N$ data points, let $\mathbf{Z} \in \mathbb{R}^{N \times d}$ denote the learned latent embeddings (where $d$ is the latent dimension, typically $d=2$ for visualization), and let $\mathbf{B} \in \mathbb{R}^{N \times p}$ denote the corresponding physical parameters (where $p$ is the number of physical parameters, e.g., $p=2$ for $(g^*, Re)$). We seek an affine transformation that maps each embedding $\mathbf{z}^{(n)} \in \mathbb{R}^d$ to its corresponding physical parameters $\mathbf{b}^{(n)} \in \mathbb{R}^p$:

\begin{equation}
\mathbf{b}^{(n)} \approx \mathbf{A} (\mathbf{z}^{(n)} - \mathbf{z}_0) + \mathbf{b}_0,
\label{eq:affine-transform}
\end{equation}

where $\mathbf{A} \in \mathbb{R}^{p \times d}$ is a linear transformation matrix, $\mathbf{z}_0 \in \mathbb{R}^d$ is a reference point in the latent space, and $\mathbf{b}_0 \in \mathbb{R}^p$ is an offset in the physical parameter space. The transformation consists of a translation (shifting by $\mathbf{z}_0$), followed by a linear mapping (multiplication by $\mathbf{A}$), and a final translation (adding $\mathbf{b}_0$).

To simplify the estimation, we can absorb the offset $\mathbf{b}_0$ into the transformation by choosing $\mathbf{z}_0$ such that $\mathbf{A} \mathbf{z}_0$ corresponds to the mean of the physical parameters. Alternatively, we can express the affine transformation in homogeneous coordinates:

\begin{equation}
\begin{bmatrix} \mathbf{b}^{(n)} \\ 1 \end{bmatrix} = \mathbf{M} \begin{bmatrix} \mathbf{z}^{(n)} \\ 1 \end{bmatrix},
\label{eq:homogeneous}
\end{equation}

where $\mathbf{M} \in \mathbb{R}^{(p+1) \times (d+1)}$ is an augmented transformation matrix that combines both the linear transformation and translation terms.

\subsection{Least-Squares Estimation}

The parameters of the affine transformation are estimated via least-squares regression. We formulate the problem as finding the matrix $\mathbf{M}$ that minimizes the squared reconstruction error:

\begin{equation}
\min_{\mathbf{M}} \sum_{n=1}^{N} \left\| \mathbf{b}^{(n)} - \mathbf{M} \begin{bmatrix} \mathbf{z}^{(n)} \\ 1 \end{bmatrix} \right\|^2.
\label{eq:least-squares}
\end{equation}

In matrix form, this can be written as:

\begin{equation}
\min_{\mathbf{M}} \| \mathbf{B} - \mathbf{Z}_{\text{aug}} \mathbf{M}^T \|_F^2,
\label{eq:matrix-form}
\end{equation}

where $\mathbf{Z}_{\text{aug}} = [\mathbf{Z}, \mathbf{1}_N] \in \mathbb{R}^{N \times (d+1)}$ is the augmented embedding matrix (with a column of ones appended), $\mathbf{1}_N$ is a column vector of $N$ ones, and $\|\cdot\|_F$ denotes the Frobenius norm.

The solution to this least-squares problem is given by:

\begin{equation}
\mathbf{M}^T = (\mathbf{Z}_{\text{aug}}^T \mathbf{Z}_{\text{aug}})^{-1} \mathbf{Z}_{\text{aug}}^T \mathbf{B},
\label{eq:lstsq-solution}
\end{equation}

which can be computed efficiently using standard linear algebra routines (e.g., \texttt{numpy.linalg.lstsq}).

Once $\mathbf{M}$ is obtained, we extract the linear transformation matrix $\mathbf{A}$ and the translation vector $\mathbf{c}$:

\begin{align}
\mathbf{A} &= \mathbf{M}_{:, 1:d} \in \mathbb{R}^{p \times d}, \\
\mathbf{c} &= \mathbf{M}_{:, d+1} \in \mathbb{R}^p,
\end{align}

where $\mathbf{M}_{:, 1:d}$ denotes the first $d$ columns of $\mathbf{M}$, and $\mathbf{M}_{:, d+1}$ denotes the $(d+1)$-th column. The reference point $\mathbf{z}_0$ in the latent space that maps to the origin in the physical parameter space is then:

\begin{equation}
\mathbf{z}_0 = -\mathbf{A}^{-1} \mathbf{c},
\label{eq:reference-point}
\end{equation}

assuming $\mathbf{A}$ is invertible (which requires $p = d$ and $\mathbf{A}$ to be full rank).

\subsection{Application to Latent Embeddings}

For each dimensionality reduction method, we independently estimate an affine transformation that maps its latent embeddings to the physical parameter space. Specifically:

\begin{enumerate}
    \item \textbf{Embedding extraction}: For each method (t-SNE, UMAP, GM-VAE), we obtain latent embeddings $\mathbf{Z}_{\text{method}} \in \mathbb{R}^{N \times 2}$.
    
    \item \textbf{Transformation estimation}: We solve the least-squares problem in \eqref{eq:matrix-form} to obtain method-specific transformation parameters $(\mathbf{A}_{\text{method}}, \mathbf{z}_{0,\text{method}})$.
    
    \item \textbf{Transformation application}: We apply the estimated transformation to align the embeddings:
    \begin{equation}
    \mathbf{Z}_{\text{method}}^{\text{trans}} = (\mathbf{Z}_{\text{method}} - \mathbf{1}_N \mathbf{z}_{0,\text{method}}^T) \mathbf{A}_{\text{method}}^T.
    \label{eq:apply-transform}
    \end{equation}
    
    \item \textbf{Visualization}: The transformed embeddings $\mathbf{Z}_{\text{method}}^{\text{trans}}$ are visualized alongside the original physical parameters $\mathbf{B}$ to assess how well each method recovers the physical structure.
\end{enumerate}

\subsection{Interpretation and Limitations}

This affine mapping approach provides a quantitative means to compare different dimensionality reduction methods by measuring how well their latent spaces align with known physical parameters. A method that produces embeddings that can be accurately mapped to physical parameters via a simple affine transformation suggests that the learned representation captures the underlying physical structure in a geometrically consistent manner.

However, several limitations should be noted:

\begin{itemize}
    \item \textbf{Post-hoc alignment}: The transformation is learned after the embeddings are obtained, meaning it does not influence the training process of the dimensionality reduction method itself. It serves purely as an evaluation and visualization tool.
    
    \item \textbf{Linearity assumption}: The affine transformation assumes a linear relationship between latent coordinates and physical parameters. If the true relationship is highly nonlinear, the mapping may introduce distortion, and the quality of the alignment may not fully reflect the method's ability to capture physical structure.
    
    \item \textbf{Scale and rotation invariance}: The transformation accounts for arbitrary rotations, scalings, and translations of the latent space. This is appropriate for comparing methods, as these transformations do not affect the intrinsic geometry of the manifold, but it means that the absolute orientation and scale of the embeddings are not preserved.
    
    \item \textbf{Overfitting risk}: When the number of data points $N$ is small relative to the number of parameters in $\mathbf{M}$ (i.e., $(d+1) \times p$), the least-squares solution may overfit to the training data. In practice, for $d=2$ and $p=2$, this requires at least $N \gg 6$ data points, which is typically satisfied in our applications.
\end{itemize}

Despite these limitations, the affine mapping provides a principled and interpretable framework for comparing dimensionality reduction methods, particularly when the goal is to assess whether learned representations align with known physical quantities. The transformation enables direct visualization of how well each method recovers the physical parameter space, facilitating both qualitative assessment (through visualization) and quantitative evaluation (through residual analysis or correlation metrics).

\section{Implementation Details of the GM-VAE Models} \label{sec:impl-details}

We adopt a U-Net architecture with modifications suitable for the Schlieren images from the laser-induced combustion experiments. The U-Net is composed of two main parts: the \texttt{UNetEncoder} and the \texttt{UNetDecoder}.

\subsection{Encoder}
The encoder section begins with an input convolutional layer (\texttt{DoubleConv}) that processes the initial input channels, followed by four down-sampling layers (\texttt{Down}) that progressively reduce the spatial dimensions while increasing the feature depth. These layers are designed to capture hierarchical features of the input data. A flattening layer converts the spatially structured data into a 1D format suitable for fully connected layers. The linear section consists of two fully connected layers that map the flattened features into a latent space representation with twice the specified latent dimension, facilitating the separation into mean and variance components for potential variational autoencoder applications.

\subsection{Decoder}
The decoder section starts by reversing the process, beginning with linear layers that map the latent space back into a high-dimensional space, followed by reshaping into the original spatial structure. It then uses four up-sampling layers (\texttt{Up}) to reconstruct the high-resolution output from the encoded features gradually. The final layer (\texttt{OutConv}) outputs a two-channel reconstruction, separating the point estimation and uncertainty channels.

\subsection{Pre-Training}
We observe that pre-training the encoder and decoder of the GM-VAE model as a standard VAE enhances the robustness of the GM-VAE embeddings by improving the initialization of GMM clusters. For experiments in Section 5.3, we pre-train the model on a subset of the dataset, including 450 training images and 75 validation images. The pre-training was efficient due to the small data size.

\subsection{Hyperparameters} \label{subsec:hyperparams}

Table~\ref{tab:hyperparams} summarizes the key hyperparameters used for training the GM-VAE model with U-Net architectures. These hyperparameters are organized into categories: architecture parameters that define the network structure, pretraining parameters for the initial VAE training phase, training parameters for the main GM-VAE optimization, and GM-VAE-specific parameters that govern the mixture model and expectation-maximization algorithm.

\begin{table}[h]
\centering
\caption{Hyperparameters for GM-VAE training with U-Net architectures.}
\label{tab:hyperparams}
\begin{tabular}{llll}
\toprule
\textbf{Category} & \textbf{Parameter} & \textbf{Combustion} & \textbf{Flow past cylinders} \\
\midrule
\multirow{3}{*}{Architecture} & U-Net channels & 8, 8, 8, 16, 16 & 8, 16, 32, 64 \\
 & Latent dimension & 2--32  & 2 \\
 & Number of clusters & 3--18  & 5 \\
\midrule
\multirow{3}{*}{Pretraining} & Learning rate & $1 \times 10^{-3}$ & $1 \times 10^{-3}$ \\
 & Weight decay & 0 & 0 \\
 & Epochs & 50 & 100 \\
\midrule
\multirow{4}{*}{Training} & Learning rate & $5 \times 10^{-4}$ & $1 \times 10^{-4}$ \\
 & Weight decay & $1 \times 10^{-4}$ & $5 \times 10^{-4}$ \\
 & Batch size & 75 & 64 \\
 & Epochs & 500 & 100 \\
\bottomrule
\end{tabular}
\end{table}

The architecture parameters control the capacity and structure of the encoder-decoder network. The U-Net convolutional channels ($c_1$ through $c_5$) determine the feature depth at each down-sampling and up-sampling stage, with values [8, 8, 8, 16, 16] used for combustion images and [8, 16, 32, 64] for the cylinder flow dataset. The latent dimension ($l_d$) and number of clusters ($K$) are varied across experiments to study their impact on reconstruction quality and interpretability.

The pretraining phase initializes the encoder and decoder as a standard VAE before GM-VAE training begins. This phase uses a higher learning rate ($1 \times 10^{-3}$) and no weight decay to quickly establish a good initialization. The main training phase then fine-tunes the model with GM-VAE objectives, using different learning rates and weight decay settings appropriate for each dataset: $5 \times 10^{-4}$ with weight decay $1 \times 10^{-4}$ for combustion images, and $1 \times 10^{-4}$ with weight decay $5 \times 10^{-4}$ for flow past cylinders.

The GM-VAE-specific parameters are common to both U-Net setups: EM regularization ($\lambda_{\text{EM}} = 1 \times 10^{-4}$) prevents cluster collapse, decoder variance ($\sigma^2_{\text{dec}} = 1.0$) is fixed to simplify the reconstruction loss, VAE KL weight ($\beta = 0.1$) balances reconstruction accuracy against KL divergence, EM steps per epoch ($N_{\text{EM}} = 2$) controls GMM update frequency, and EM initialization uses k-means clustering.

For the 1D surface reaction time series dataset, we use a fully connected (linear) architecture that differs significantly from the U-Net setups. The encoder uses hidden dimensions [32, 16, 8] with $\tanh$ activations, while the decoder reverses this structure to [8, 16, 32]. This linear architecture operates on flattened time series vectors of dimension 50. Training uses 1024 training samples, 128 validation samples, and 128 test samples, with a batch size of 64. Unlike the U-Net setups, the linear architecture does not use a separate pretraining phase; instead, it trains end-to-end from random initialization. The main training phase uses a learning rate of $1 \times 10^{-3}$ with no weight decay, and runs for 20,000 epochs to allow the simpler architecture to converge. The GM-VAE-specific parameters differ as follows: decoder variance is set to $\sigma^2_{\text{dec}} = 1 \times 10^{-5}$ to encourage better reconstruction performance on the 1D time series data, EM steps per epoch is $n_{\text{EM}} = 1$ (one EM update per epoch), and cluster means are initialized randomly in $[-1, 1]$.





\bibliographystyle{unsrtnat}
\bibliography{main_CMAME}

\appendix


    



\end{document}